\documentclass[10pt,twocolumn,letterpaper]{article}

\usepackage{iccv}              
\usepackage{multirow}
\usepackage{algorithmic}
\usepackage{algorithm}
\usepackage[table]{xcolor} %

\newcommand{\colorfirst}{255, 153, 153}
\newcommand{\colorsecond}{255, 204, 153}
\newcommand{\colorthird}{255, 255, 153}
\definecolor{colorfirst}{RGB}{255, 153, 153}
\definecolor{colorsecond}{RGB}{255, 204, 153}
\definecolor{colorthird}{RGB}{255, 255, 153}

\definecolor{colorfirst}{RGB}{255, 153, 153}
\definecolor{colorsecond}{RGB}{255, 204, 153}
\definecolor{colorthird}{RGB}{255, 255, 153}

\newcommand{\teaser}{
\vspace{-2em}
\centering
\includegraphics[width=0.95\textwidth,trim=0em 0em 0em 0em,
clip]{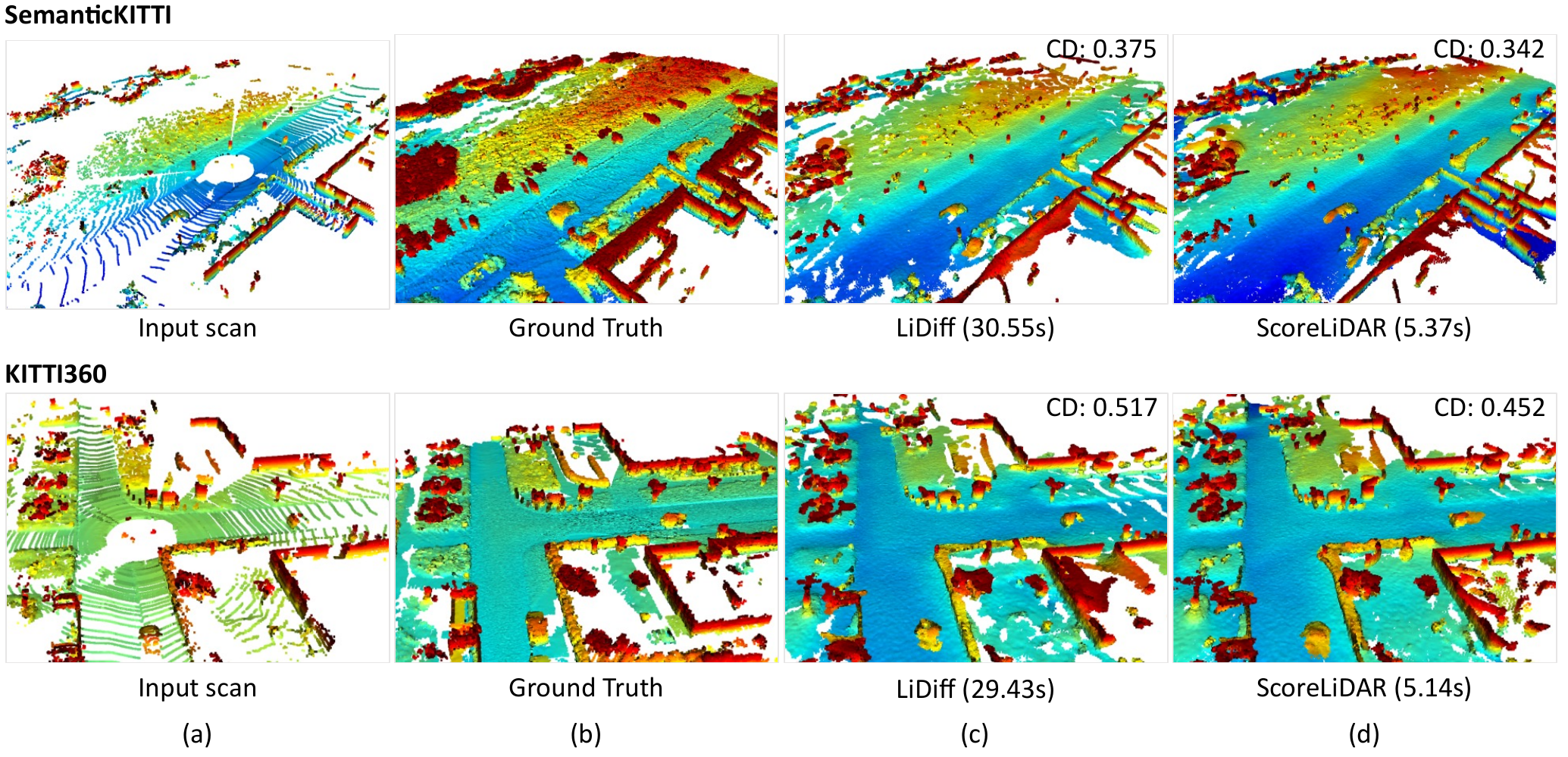}
\vspace{-1.1em}
\captionof{figure}{
A demonstration of the LiDAR scene completion examples. Given a sparse LiDAR scan in (a), the model aims to recover the ground-truth dense scene as in (b). In these examples, scans are from SemanticKITTI~\cite{semantickitti} and KITTI360~\cite{kitti360} dataset. In both cases, LiDiff~\cite{LiDiff}, a SOTA LiDAR scene completion method, requires about 30 seconds as in (c). In comparison, our proposed ScoreLiDAR takes only about 5 seconds in (d), achieving over 5x speedup with improved completion quality indicated by lower Chamfer Distance (CD).
}
\label{fig:teaser}
\vspace{1em}
}

\definecolor{iccvblue}{rgb}{0.21,0.49,0.74}
\usepackage[pagebackref,breaklinks,colorlinks,allcolors=iccvblue]{hyperref}

\def\paperID{7053} 
\def\confName{ICCV}
\def\confYear{2025}

\title{Distilling Diffusion Models to Efficient 3D LiDAR Scene Completion}

\author{Shengyuan Zhang$^1$
\quad An Zhao$^1$ \quad Ling Yang$^3$ \quad Zejian Li$^{2,*}$ \quad Chenye Meng$^1$ \\ Haoran Xu$^4$ \quad Tianrun Chen$^1$ \quad AnYang Wei$^4$ \quad Perry Pengyun GU$^4$ \quad Lingyun Sun$^1$ \\
{\small $^1$ College of Computer Science and Technology, Zhejiang University \quad \small $^2$ School of Software Technology, Zhejiang University} \\ 
{\small $^3$ Peking University \quad \small $^4$ Zhejiang Green Zhixing Technology co., ltd} \\
{\tt\small $^{1,2}$\{zhangshengyuan,zhaoan040113,zejianlee,mengcy,tianrun.chen,sunly\}@zju.edu.cn} \\
{\tt\small$^3$\{yangling0818\}@163.com }
{\tt\small$^4$\{Haoran.Xu5,weianyang,gupengyun\}@geely.com }
{\tt\small$^*$Corresponding author}
}

\begin{document}
\twocolumn[{
\maketitle
\teaser
}]
\begin{abstract}
Diffusion models have been applied to 3D LiDAR scene completion due to their strong training stability and high completion quality.
However, the slow sampling speed limits the practical application of diffusion-based scene completion models since autonomous vehicles require an efficient perception of surrounding environments. 
This paper proposes a novel distillation method tailored for 3D LiDAR scene completion models, dubbed \textbf{ScoreLiDAR}, which achieves efficient yet high-quality scene completion.
ScoreLiDAR enables the distilled model to sample in significantly fewer steps after distillation.
To improve completion quality, we also introduce a novel \textbf{Structural Loss}, which encourages the distilled model to capture the geometric structure of the 3D LiDAR scene.
The loss contains a scene-wise term constraining the holistic structure and a point-wise term constraining the key landmark points and their relative configuration.
Extensive experiments demonstrate that ScoreLiDAR significantly accelerates the completion time from 30.55 to 5.37 seconds per frame ($>$5$\times$) on SemanticKITTI and achieves superior performance compared to state-of-the-art 3D LiDAR scene completion models. Our model and code are publicly available on \url{https://github.com/happyw1nd/ScoreLiDAR}.

\end{abstract}

\section{Introduction}
\label{sec:intro}

Recognizing the surrounding environment accurately and efficiently using onboard sensors is crucial for the safe operation of autonomous vehicles~\cite{voxformer,emergent}. Among different types of sensors, 3D LiDAR has become one of the most widely adopted sensors due to its broader detection range and higher detection precision~\cite{LiDiff,meydani2023state}. However, driving scenarios are often complex, and the 3D point clouds collected by LiDAR are typically sparse, particularly in occluded areas~\cite{MID,lode}. This sparsity causes a decline in the ability to understand 3D scenes~\cite{LiDiff,diffssc}. Thus, inferring and completing sparse 3D LiDAR scenes is necessary to provide a dense and more comprehensive scene representation.

\begin{figure}[t]
    \centering
    \includegraphics[width=0.9\linewidth]{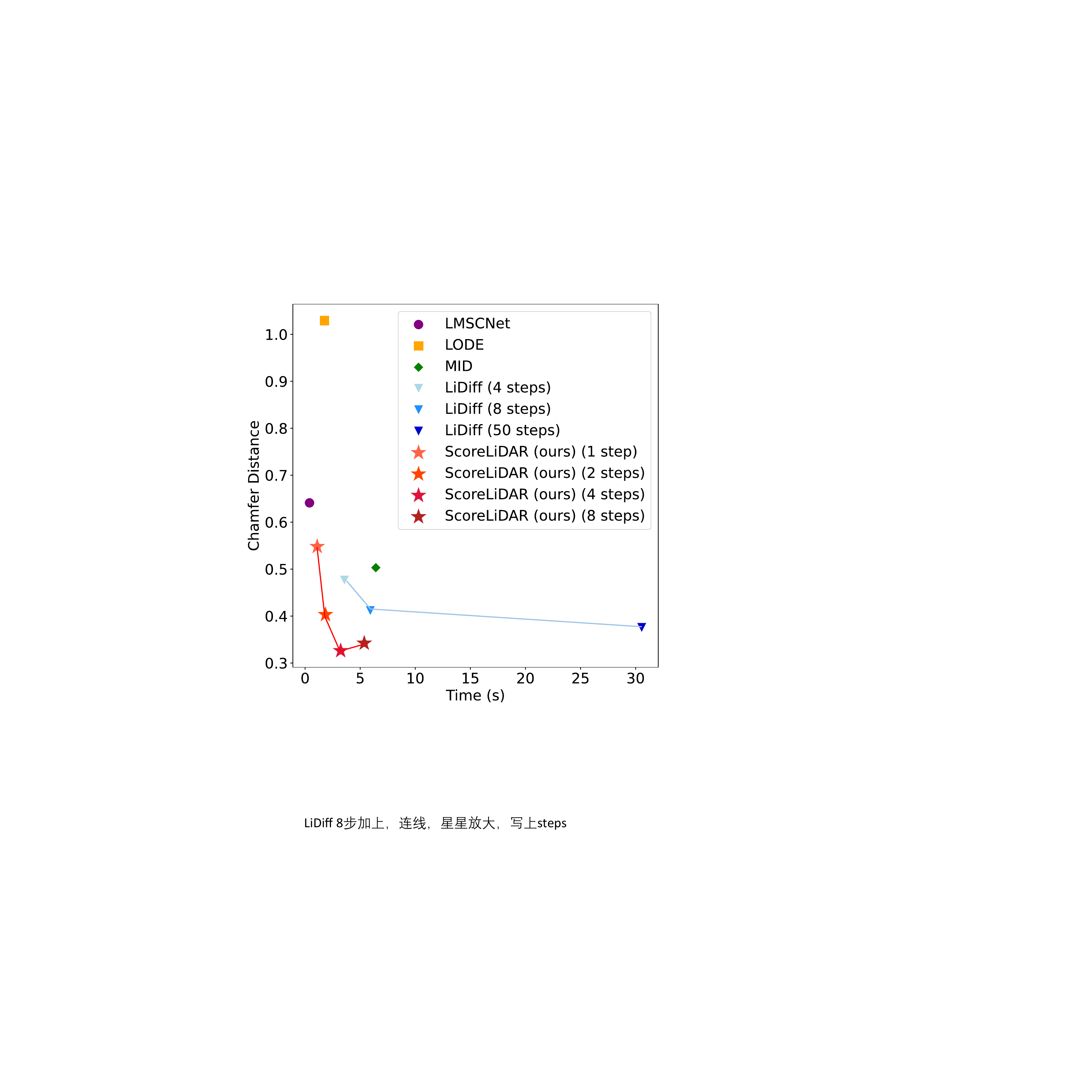}
    \vspace{-0.5em}
    \caption{A visualization of LiDAR scene completion performances with different models on SemanticKITTI~\cite{semantickitti} dataset. 
    Generally, our proposed ScoreLiDAR achieves better scene completion performance and speed trade-off.}
    \vspace{-1em}
    \label{fig:CD_time}
\end{figure}

Due to the advantages of strong training stability and high-generation quality, existing works utilize diffusion models to complete the 3D LiDAR scenes and achieve outstanding results~\cite{LiDiff,R2DM}. However, the diffusion model often requires multiple network iterations to obtain a dense, complete, and high-quality LiDAR scene, which is time-consuming~\cite{zhang2024distribution,CTM}. Autonomous vehicles require fast and efficient perception and recognition of surrounding environments, so the slow sampling speed limits the practical application of diffusion models. Although existing works have proposed acceleration methods for diffusion models~\cite{consistency_model,progressive_distillation,Diff-Instruct,yang2024consistency,wang2024rectified}, due to the differences between 3D LiDAR scenes and image data—where LiDAR scenes often contain complex geometric structure information—these techniques have not been  explored in the acceleration of diffusion-based LiDAR scene completion model.




In this work, we propose \textbf{ScoreLiDAR}, a novel distillation method tailored for 3D LiDAR scene completion diffusion models, which enables efficient and high-quality scene completion (\cref{fig:teaser} and~\cref{fig:CD_time}).
ScoreLiDAR aims to tackle the unique 3D distribution alignment challenge in LiDAR scene completion. By exploiting a bidirectional gradient guidance mechanism, it allows the student model’s completion results under sparse point cloud conditions to progressively approach the multi-step iterative reconstruction quality of the teacher model. 
Given a completed scene generated by the student model, the teacher model predicts the scene's score as a gradient to enhance realism, while a designed auxiliary model predicts the scene's another score as a gradient to suppress unrealistic completion. 
The student model is then updated according to the difference between these two gradients. This update drives the completion process toward more structurally coherent LiDAR scenes. Finally, we introduce a \textbf{Structural Loss} consisting of a scene-wise term and a point-wise term constraining the key landmark points and their relative configuration. Prior studies~\cite{prolificdreamer,wang2024taming,ma2024scaledreamer,luciddreamer} demonstrated that the bidirectional gradient guidance mechanism can effectively accelerate 3D rendering speed. In this work, we argue that a similar approach can be directly applied to the distillation of diffusion-based LiDAR scene completion models.



Our contribution can be summarized as follows: (1) We propose \textbf{ScoreLiDAR}, a novel distillation method tailored for diffusion-based 3D LiDAR scene completion models, which achieves efficient scene completion. (2) We introduce a \textbf{Structural Loss} to effectively capture the geometric structure information of 3D point clouds during the distillation process, which ensures high-quality scene completion. (3) Extensive experiments show that ScoreLiDAR enables fast and efficient scene completion while achieving optimal generation quality compared to the existing models. 
\section{Related work}
\paragraph{3D LiDAR Scene completion} 
3D LiDAR scene completion refers to recovering a complete scene from a sparse, incomplete LiDAR scan in applications such as autonomous driving~\cite{completion1,MID}. Current mainstream LiDAR scene completion methods include depth completion-based and Signed Distance Field (SDF)-based approaches. Depth completion-based methods aim to recover dense depth maps from sparse depth measurements~\cite{depth1,depth2,depth3}. These methods typically leverage deep learning techniques~\cite{depth4,depth5} and can also incorporate guidance from RGB images to achieve higher-quality completion results~\cite{depth6,depth7,depth8}. SDF-based methods represent scenes as voxel grids, with the core idea of using signed distance fields to complete sparse LiDAR scenes~\cite{MID,lode}. These methods are constrained by voxel resolution, making them prone to losing details within the scene~\cite{LiDiff,dai2018scancomplete}. In addition, some methods introduce semantic information to enhance LiDAR scene completion~\cite{roldao20223d,scpnet}. These methods can generate dense and complete scenes while providing semantic labels for each point, leading to broader application potential~\cite{bissc,wang2023semantic}.

\paragraph{Diffusion-based 3D LiDAR scene completion}
\label{sec:rw}
Due to the strong training stability and high generation quality of diffusion models, many methods leverage diffusion models for LiDAR scene completion tasks~\cite{LiDiff,diffssc,R2DM,LiDMs,lee2023diffusion}. The work of Lee~\textit{et al}.~\cite{lee2023diffusion} is the first to apply diffusion models at the scene scale for LiDAR scene completion, enabling the generation of realistic scenes conditioned on partial observations from sparse point clouds. Similarly, R2DM~\cite{R2DM} utilizes diffusion models based on distance and reflectance intensity image representations to generate various high-fidelity 3D LiDAR scenes. LiDiff~\cite{LiDiff} indicates that adding noise to point cloud data at the scene scale leads to a loss of detail. Therefore, LiDiff proposes operating directly on individual points and redefines the noise schedule and denoising processes to generate scenes with richer detail. Based on LiDiff, DiffSSC~\cite{diffssc} further performs semantic scene completion tasks by implementing denoising and segmentation separately in both the point and semantic spaces. Moreover, LiDMs~\cite{LiDMs} constructs the pipeline from the perspectives of pattern realism, geometric realism, and object realism, achieving generation under different conditions. 


Due to the slow sampling inherent in diffusion models, diffusion-based 3D LiDAR scene completion suffers from slow inference, hindering its application in autonomous vehicles. Thus, this paper proposes a distillation method for diffusion-based 3D LiDAR scene completion models to achieve faster and higher-quality completion.

\section{Preliminary}
\label{sec:preli}

\subsection{Brief introduction of diffusion models}
\label{subsec:dm}
The diffusion models have two processes: forward diffusion and reverse denoising process~\cite{DDPM,DDIM}. In the forward diffusion process, given the data $\boldsymbol{x}^0\sim q(\boldsymbol{x})$ from the training distribution, the diffusion model adds different scales of noise to $\boldsymbol{x}^0$ according to different timesteps $t\in [1,T]$ to obtain noisy data $\{\boldsymbol{x}^1,\boldsymbol{x}^2,\ldots,\boldsymbol{x}^T \}$. When $T$ is large enough, $\boldsymbol{x}^T$ approaches to standard Gaussian distribution, namely, $q(\boldsymbol{x}^T)\approx \mathcal{N}(0,I)$. This process is parameterized by a series of predefined noise factors $\beta^t$. By defining $\alpha^t = 1- \beta^t$, the diffusion process is expressed as~\cite{DDPM}:
\begin{equation}
\label{eq:forward_process}
    \boldsymbol{x}^t = \sqrt{\bar{\alpha}^t} \boldsymbol{x}^0 + \sqrt{1-\bar{\alpha}^t}\boldsymbol{\epsilon}^t
\end{equation}
Here $\bar{\alpha}^t = \prod_{t=1}^{T}\alpha^t$, $p(\boldsymbol{x}^t\mid \boldsymbol{x}^0)= \mathcal{N} (\sqrt{\bar{\alpha}^t} , (1-\bar{\alpha}^t)I )
$. 

During the training, the diffusion model tries to predict the added noise at different timesteps $t$. Given the input $\boldsymbol{x}^0$ and the condition $c$ (optional), the noisy data $\boldsymbol{x}^t$ can be calculated by~\cref{eq:forward_process}. The diffusion model $\boldsymbol{\epsilon}_\theta$ predicts the noise according to $\boldsymbol{x}^t, c, t$ and is then optimized by calculating the $\ell_2$ loss between the predicted and the real noise. 
\begin{equation}
\label{eq:dm_loss}
    \mathcal{L}_{DM} = \mathbb{E}_{t, \epsilon}\left[\|\boldsymbol{\epsilon} - \boldsymbol{\epsilon}_\theta(\boldsymbol{x}^t,c,t)\|^2\right]
\end{equation}
Here $\theta$ is the trainable parameter of $\boldsymbol{\epsilon}_\theta$.

In the reverse denoising process, the diffusion model starts from the timestep $T$ and progressively removes the predicted noise to obtain a generated sample. The process of denoising $\boldsymbol{x}^t$ to obtain $\boldsymbol{x}^{t-1}$ can be written as in~\cite{DDPM}
\begin{equation}
\label{eq:reverse_process}
    \boldsymbol{x}^{t-1}=\frac{1}{\sqrt{\alpha^{t}}}\left(\boldsymbol{x}^{t}-\frac{1-\alpha^{t}}{\sqrt{1-\bar{\alpha}^{t}}} \boldsymbol{\epsilon}_{\theta}\left(\boldsymbol{x}^{t}, c, t\right)\right)+\sigma^{t} \textbf{z}
\end{equation}
Here $\textbf{z} \sim \mathcal{N}(0,I)$. In this process, the number of required inference steps varies depending on different sampling methods. For instance, DDPM~\cite{DDPM} requires 1000 timesteps, while DDIM~\cite{DDIM} and DPM solver~\cite{dpm} can reduce this to no more than 100 timesteps.

\subsection{3D LiDAR scene completion diffusion models}
\label{subsec:lidar}
The 3D LiDAR scene completion diffusion models take the incomplete scan $\mathcal{P}=\{\boldsymbol{p}_1,\boldsymbol{p}_2,...,\boldsymbol{p}_N\}$ ( $p_i\in\mathbb{R}^3$) and try to generate the complete scene $\mathcal{G}^0= \{\boldsymbol{p}^0_1,\boldsymbol{p}^0_2,...,\boldsymbol{p}^0_M\}$.
Given the input LiDAR scan $\mathcal{P}$ and ground truth $\mathcal{G}$, a diffusion model can be trained to perform 3D LiDAR scene completion. The noisy scene $\mathcal{G}^t$ at timestep $t$ is calculated from the ground truth $\mathcal{G}$ at point level~\cite{LiDiff, diffssc},
\begin{equation}
\label{eq:forward_diffuison_lidar}
    \boldsymbol{p}_{m}^{t}=\sqrt{\bar{\alpha}^{t}} \boldsymbol{p}_{m}+\sqrt{1-\bar{\alpha}^{t}} \epsilon^t, \forall \boldsymbol{p}_{m} \in \mathcal{G}
\end{equation}
Here $\mathcal{G}^t=\{\boldsymbol{p}^t_1,\boldsymbol{p}^t_2,...,\boldsymbol{p}^t_M\}$. Because the LiDAR point cloud is sparse, the noisy data retains very little information about the original data. To generate more realistic point cloud scenes, the LiDAR scan $\mathcal{P}$ can be used as a condition of the diffusion model~\cite{LiDiff}. In this case, the training loss of the diffusion model is given by:
\begin{equation}
\label{eq:lidar_dm_loss}
    \mathcal{L}_{DM}=\mathbb{E}_{t, \epsilon} \left[\left\|\boldsymbol{\epsilon}-\boldsymbol{\epsilon}_{\theta}\left(\mathcal{G}^{t}, \mathcal{P}, t\right)\right\|^{2}\right]
\end{equation}

Then, as described in~\cref{subsec:dm}, the completed scene ${\mathcal{G}}^0$ can be generated by progressive denoising from $\mathcal{G}^T$. Because the scale of the LiDAR scene is large and the data range is different across different point cloud axes, directly normalizing the entire scene compresses the data into a smaller range, which potentially leads to the loss of critical details~\cite{LiDiff,diffssc}. To solve this issue, LiDiff~\cite{LiDiff} modifies the diffusion process by adding a local noise offset to each point $\boldsymbol{p}_m$, gradually perturbing the point cloud at each timestep. For~\cref{eq:forward_process}, $\boldsymbol{x}^0$ is set to 0, and $\boldsymbol{x}^t$ is added to each point $\boldsymbol{p}_m$,
\begin{equation}
\label{eq:lidar_point_noisy}
    \boldsymbol{p}_{m}^{t} =\boldsymbol{p}_{m}+\left(\sqrt{\bar{\alpha}^{t}} \mathbf{0}+\sqrt{1-\bar{\alpha}^{t}} \boldsymbol{\epsilon^t}\right) =\boldsymbol{p}_{m}+\sqrt{1-\bar{\alpha}^{t}} \boldsymbol{\epsilon^t}
\end{equation}

\begin{figure*}[!t]
    \centering
    \includegraphics[width=0.95\linewidth]{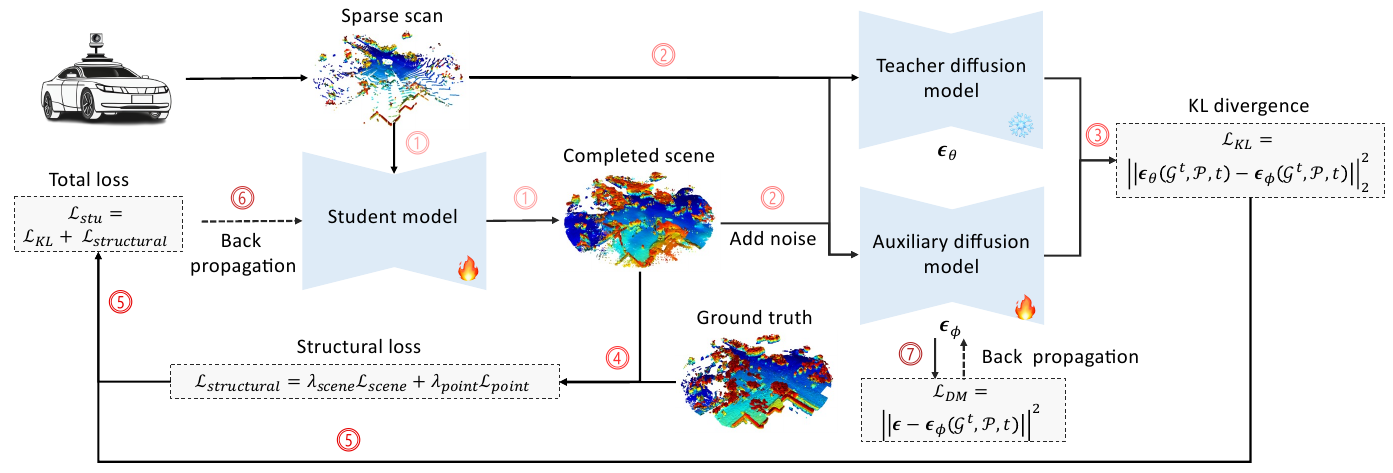}
    \caption{The overall structure of ScoreLiDAR. (1) The student model generates the completed scene based on the sparse scan. (2) The sparse scan and noisy completed scene are input to $\boldsymbol{\epsilon}_\theta$ and $\boldsymbol{\epsilon}_\phi$. (3) The predicted score of $\boldsymbol{\epsilon}_\theta$ and $\boldsymbol{\epsilon}_\phi$ are used to calculated the KL divergence. (4) Structural loss is calculated based on the completed scene and the ground truth. (5) The total loss is calculated with KL divergence and structural loss. (6) The student model is optimized according to the total loss. (7) The diffusion model $\boldsymbol{\epsilon}_\phi$ is updated with the completed scene.}
    \vspace{-1em}
    \label{fig:structure}
\end{figure*}

Due to this special case, the initial noisy scene $\mathcal{G}^{T}$ cannot directly start from standard Gaussian noise in the sampling process. Instead, the LiDAR scan $\mathcal{P}$ is used to obtain $\mathcal{G}^{T}$~\cite{LiDiff}. 
Firstly, given the initial incomplete scan $\mathcal{P}$, the number of the point clouds is increased by duplicating the original points $K$ times and getting the pseudo dense scan $\mathcal{P}^*=\{\boldsymbol{p}^*_1,\boldsymbol{p}^*_2, \ldots ,\boldsymbol{p}^*_{M}\}$, where we assume $M=KN$. 
Then, we calculate the noisy point cloud $\mathcal{P}^T$ by~\cref{eq:lidar_point_noisy}. As $\mathcal{P}^T$ is noisy enough, it can be regarded as $\mathcal{G}^T$ during the training. After that, a step-by-step denoising process is applied to obtain the completed scene $\mathcal{G}^0$.
\section{Method}
Our goal is to distill a pre-trained 3D LiDAR scene completion diffusion model into a student model with significantly fewer sampling steps, enabling efficient and high-quality scene completion. Firstly, we introduce the distillation method tailored for 3D LiDAR scene completion diffusion models in~\cref{subsec:score_disillation}. Then, we introduce the structural loss to improve the distillation process with both scene-wise loss and point-wise loss in~\cref{subsec:structural_loss}. Finally, we describe the optimization procedure of ScoreLiDAR in~\cref{subsec:optimization}. The overall structure of ScoreLiDAR is shown in~\cref{fig:structure}.

\subsection{Distillation for 3D LiDAR scene completion}
\label{subsec:score_disillation}
Ideally, the final student model would achieve completion results comparable to those of the teacher model at a faster speed. In the 3D LiDAR scene completion scenario, let $q^0$ be the distribution of the ground truth $\mathcal{G}$, and $\boldsymbol{\epsilon}_\theta$ be the pre-trained scene completion diffusion model whose multi-step generated distribution approximates $q^0$. Let $G_\mathit{stu}$ be the student model that can perform efficient LiDAR scene completion with the generated distribution $p_G^0$. ScoreLiDAR aims to minimize the KL divergence between the distribution of the teacher model and the generated distribution of the student model~\cite{prolificdreamer,DMD}.
\begin{equation}
\label{eq:KL}
    \min _{\eta} D_{K L}\left(p^{0}_{G}\left(\mathcal{G}^0; \eta \right) \| q^{0}\left(\mathcal{G}^0\right)\right)
\end{equation}
Here $\mathcal{G}^0$ is the completed scene generated by one-step sampling of $G_\mathit{stu}$ conditioned on $\mathcal{P}$, for simplicity, we omit $\mathcal{P}$ when representing the distribution, $\eta$ is the trainable parameter of $G_\mathit{stu}$. However, the high-density regions of $q^0$ are sparse in the data space, so it is hard to directly solve~\cref{eq:KL}. According to Theorem 1 in~\cite{prolificdreamer}, we expand the optimization problems in~\cref{eq:KL} by minimizing the KL divergence between two distributions at different noise levels as:
\begin{equation}
\label{eq:KL_extend}
    \min _{\eta} \mathcal{L}_{KL} = \mathbb{E}_{t, \epsilon}\left[ D_{K L}\left(p^{t}_{G}\left(\mathcal{G}^t\right) \| q^{t}\left(\mathcal{G}^t\right)\right)\right]
\end{equation}
Here $t$ is the timestep controlling the noise level, $\epsilon$ is random noise, and $\mathcal{G}^t=\{\boldsymbol{p}^t_1,\boldsymbol{p}^t_2,\ldots,\boldsymbol{p}^t_M\}$ is the noisy version of the completed scene $\mathcal{G}^0$ at timestep $t$. The gradient of $G_\mathit{stu}$ in~\cref{eq:KL_extend} is approximated by
\begin{equation}
\label{eq:gradient}
\begin{aligned}
    & \nabla_\eta D_{\mathrm{KL}} \left(p_G^{t} \left( \mathcal{G}^t \right) \| q^{t}\left(\mathcal{G}^t\right)\right) \\ & = 
    \mathbb{E}_{t,\epsilon} \left[ \nabla_{\mathcal{G}^t}\log p_G^{t}\left(\mathcal{G}^t\right) - \nabla_{\mathcal{G}^t} \log q^{t}\left(\mathcal{G}^t\right)\right] \frac{\partial \mathcal{G}^t}{\partial \eta}
\end{aligned}
\end{equation}

We use the pre-trained diffusion model $\boldsymbol{\epsilon}_\theta$ to approximate $-\sqrt{1-\bar{\alpha}^t}\nabla_{\mathcal{G}^t} \log q^{t}\left(\mathcal{G}^t\right)$ as discussed in Section 5 of Supplementary Materials (Sec S5).
Similarly, $-\sqrt{1-\bar{\alpha}^t}\nabla_{\mathcal{G}^t}\log p_G^{t}\left(\mathcal{G}^t\right)$ is approximated by an auxiliary diffusion model $\boldsymbol{\epsilon}_\phi$, which is independently trained with the denoising loss \cref{eq:lidar_dm_loss} on generated samples $\mathcal{G}^0$. Then, with the simplification as in~\cite{DDPM,prolificdreamer}, the $\mathcal{L}_{KL}$ is estimated by
\begin{equation}
\label{eq:G_loss}
    \mathcal{L}_{KL} \approx \mathbb{E}_{t, \epsilon} \left[\|\boldsymbol{\epsilon}_{\theta}\left(\mathcal{G}^{t}, \mathcal{P}, t\right)- \boldsymbol{\epsilon}_{\phi}\left(\mathcal{G}^{t}, \mathcal{P}, t\right)\|^2_2\right]
\end{equation}
Thus, the gradient in~\cref{eq:gradient} is approximated by
\begin{equation}
\label{eq:G_gradient}
\begin{aligned}
    & \nabla_\eta D_{\mathrm{KL}} \left(p_G^{t} \left( \mathcal{G}^t \right) \| q^{t}\left(\mathcal{G}^t\right)\right) \\ & \approx 
    \mathbb{E}_{t,\epsilon}\left[ \boldsymbol{\epsilon}_{\theta}\left(\mathcal{G}^{t}, \mathcal{P}, t\right)- \boldsymbol{\epsilon}_{\phi}\left(\mathcal{G}^{t}, \mathcal{P}, t\right)\right] \frac{\partial \mathcal{G}^t}{\partial \eta} 
\end{aligned}
\end{equation}
Intuitively, the orientation of $-\boldsymbol{\epsilon}_\theta(\mathcal{G}^{t}, \mathcal{P}, t)$ points to the pre-trained distribution, while that of $-\boldsymbol{\epsilon}_\phi(\mathcal{G}^{t}, \mathcal{P}, t)$ points to the student model's generative distribution. Thus, decenting along the bidirectional gradient $\left[ \boldsymbol{\epsilon}_{\theta}\left(\mathcal{G}^{t}, \mathcal{P}, t\right)- \boldsymbol{\epsilon}_{\phi}\left(\mathcal{G}^{t}, \mathcal{P}, t\right)\right]$ updates the student model’s generative distribution toward the pre-trained distribution, achieving a more accurate completion. 
Sec S2.1 discusses the efficiency of the distillation process.


\subsection{Structural loss}
\label{subsec:structural_loss}
Although the distillation process in~\cref{subsec:score_disillation} is effective in training models~\cite{Diff-Instruct,zhang2024distribution}, we found that directly applying it to LiDAR scene completion diffusion models leads to loss of local details and reduced realism. This is because the point cloud in LiDAR scenes includes complex geometric information that is not explicitly captured by diffusion models. Thus, we introduce a structural loss to further refine the distillation process and improve the completion quality. This structural loss includes scene-wise loss and point-wise loss and can help the student model effectively capture geometric structure information of the 3D point clouds.

\paragraph{Scene-wise loss.} In the distillation process mentioned in~\cref{subsec:score_disillation}, the gradient $\nabla_\eta D_{\mathrm{KL}}$ in~\cref{eq:G_gradient} is well-defined when $t\gg 0$, \textit{i.e.} the generated samples are totally disturbed by Gaussian noise. However, $\nabla_\eta D_{\mathrm{KL}}$ becomes unreliable when $t$ is small~\cite{zhang2024distribution,DMD}. This is because the student model often generates subpar results at the early stage due to the complexity of the point cloud data. It is easy for the noisy generated samples to lie outside the training distribution of the teacher model, causing the unreliable network prediction of the teacher model~\cite{zhang2024distribution,DMD}.

To solve this issue, we introduce the scene-wise loss, which minimizes the distance between the ground truth scene $\mathcal{G}$ and the completed scene $\mathcal{G}^0$,
\begin{equation}
    \label{eq:scene_wise}
    \mathcal{L}_{scene} = \frac{1}{|\mathcal{G}^0|} \sum_{\boldsymbol{p}^0_i \in \mathcal{G}^0} \min _{\boldsymbol{p} \in \mathcal{G}}\|\boldsymbol{p}^0_i-\boldsymbol{p}\|^{2}
\end{equation}
This loss calculates the mean squared error between each point $\boldsymbol{p}^0_i$ in the generated scene $\mathcal{G}^0$ and its closest corresponding point $\boldsymbol{p}$ in the ground truth $\mathcal{G}$. It helps the student model capture the holistic structure, which prevents the optimization direction from deviating in the early stages and enhances training stability. The scene-wise loss enables the generated scenes to be closer to the ground truth globally, thereby enhancing the completion quality and fidelity.

\paragraph{Point-wise loss.} 
As seen in~\cref{eq:G_gradient}, the distillation process only constrains the overall distribution of the completed scene, ignoring the relative positions between different points. Directly using the gradient in~\cref{eq:G_gradient} to optimize the student model may lead to loss of local details.

Thus, we introduce the point-wise loss to capture the relative structural information between different points in the 3D LiDAR scene. The point-wise loss calculates the difference between the inter-point distance matrices of the completed scene and the ground truth. Due to the large number of points in the scene, calculating the distance matrix for all points is computationally intensive. Therefore, we select $n$ key points to compute the distance matrix with $n \ll M$. Based on the local geometric features of each point, we choose key points that are critical for representing the structure of the 3D LiDAR scene. For each point $\boldsymbol{p}^0_i$ in the completed scene $\mathcal{G}^0$, we find its K-nearest neighbor, denoted as the set $\mathcal{K}_i$. Then we select the key points by calculating their curvature $\kappa_i$. The specific steps are as follows:
\begin{itemize}
    \item Calculate the centroid $\bar{\boldsymbol{p}}^0_i$ of the neighborhood $\mathcal{K}_i$
    \begin{equation}
        \bar{\boldsymbol{p}}^0_i = \frac{1}{K} \sum_{\boldsymbol{p}^0_j \in \mathcal{K}_i} \boldsymbol{p}^0_j
    \end{equation}
    \item Calculate the neighborhood covariance matrix  $\mathcal{C}_i$  for $\bar{\boldsymbol{p}}^0_i$
    \begin{equation}
        \mathcal{C}_i = \frac{1}{K} \sum_{\boldsymbol{p}^0_j \in \mathcal{K}_i} (\boldsymbol{p}^0_j - \bar{\boldsymbol{p}}^0_i)(\boldsymbol{p}^0_j - \bar{\boldsymbol{p}}^0_i)^T
    \end{equation}
    \item Perform eigen-decomposition on the covariance matrix $\mathcal{C}_i$ to obtain the eigenvalues $\lambda_1 < \lambda_2 <\cdots < \lambda_m$.
    \item Curvature $\kappa_i$ can be calculated using the eigenvalues
    \begin{equation}
        \kappa_i = \frac{\lambda_1}{\sum_{j=1}^m \lambda_j}
    \end{equation}
\end{itemize}
A larger curvature $\kappa_i$ indicates greater local shape variation.
Those points with great local variation are typically located at corners, edges, or endpoints, which tend to shape the main structure of the scene. Therefore, the top $n$ points with the highest curvature values are selected as key points.

Given the ground truth $\mathcal{G}= \{\boldsymbol{p}_1,\boldsymbol{p}_2,\ldots,\boldsymbol{p}_M\}$, we select $n$ key points ($n \ll M$) from the point cloud to construct the $n\times n$ distance matrix $\mathcal{D}$. The $d_{ij} \in \mathcal{D}$ represents the pairwise Euclidean Distance between points $i$ and $j$. Then, for completed scene $\mathcal{G}^0$, we select $n$ points that are closest to the key points in $\mathcal{G}$ as the corresponding key points to obtain the distance matrix $\mathcal{D}_G$. Thus, the point-wise loss is calculated by
\begin{equation}
\label{eq:inner_loss}
\mathcal{L}_{point} = \mathbb{E}_{t, \epsilon} \left[\|\mathcal{D} - \mathcal{D}_G\|^2_2\right]
\end{equation}

The point-wise loss can help the student model capture the relative configuration of key points and further enhance the geometric accuracy and detail retention of the completed scene. This ensures that key objects like cars, traffic cones, and walls are better completed, which is crucial for autonomous vehicles to recognize surroundings accurately.

\paragraph{Structural loss.} The final structural loss of $G_\mathit{stu}$ is 
\begin{equation}
    \mathcal{L}_{structural} = \lambda_{scene}\mathcal{L}_{scene} + \lambda_{point}\mathcal{L}_{point}
\end{equation}
Here $\lambda_{scene}$ and $\lambda_{point}$ are the weight of scene-wise loss and point-wise loss.


\subsection{Optimization procedure}
\label{subsec:optimization}
During the training, $G_\mathit{stu}$ and $\boldsymbol{\epsilon}_\phi$ are initialized with the teacher model $\epsilon_\theta$ and optimized alternately. The auxiliary diffusion model $\boldsymbol{\epsilon}_\phi$ is trained on the completed scene of the student model with~\cref{eq:lidar_dm_loss}. As for $G_\mathit{stu}$, we follow the proposed method to select $\frac{1}{30}$ of the points from the entire point cloud as key points for calculating the point distance matrix. Then, $G_\mathit{stu}$ is optimized with the following objective
\begin{equation}
\label{eq:objective}
\mathcal{L}_{stu} = \mathcal{L}_{KL} + \mathcal{L}_{structural}
\end{equation}
We set $\lambda_{scene}=0.5$ and $\lambda_{point}=0.01$ defaultly. The implementation details are provided in Sec S1. 

\label{sec:method}
\begin{table}[t]
\centering
\resizebox{\columnwidth}{!}{%
\begin{tabular}{lrrrr}
\toprule
Model & \multicolumn{1}{c}{CD $\downarrow$} & \multicolumn{1}{c}{JSD $\downarrow$} & \multicolumn{1}{c}{EMD $\downarrow$} & \multicolumn{1}{c}{Times (s) $\downarrow$} \\ \midrule
LMSCNet~\cite{lmscnet} & 0.641 & 0.431 & - & \cellcolor[RGB]{\colorfirst}0.40 \\
LODE~\cite{lode} & 1.029 & 0.451 & - & \cellcolor[RGB]{\colorsecond}1.76 \\
MID~\cite{MID} & 0.503 & 0.470 & - & 6.42 \\
PVD~\cite{pvd} & 1.256 & 0.498 & - & 262.54 \\
LiDiff~\cite{LiDiff} & 0.434 & 0.444 & \cellcolor[RGB]{\colorfirst}22.15 & 30.38 \\
LiDiff (Refined)~\cite{LiDiff} &  \cellcolor[RGB]{\colorsecond}0.375 &  \cellcolor[RGB]{\colorsecond}0.416 & \cellcolor[RGB]{\colorthird}23.16 & 30.55 \\ 
\midrule
ScoreLiDAR & \cellcolor[RGB]{\colorthird}0.406 & \cellcolor[RGB]{\colorthird}0.425 & \cellcolor[RGB]{\colorsecond}23.14 & \cellcolor[RGB]{\colorthird}5.16 \\
ScoreLiDAR (Refined) & \cellcolor[RGB]{\colorfirst}0.342 & \cellcolor[RGB]{\colorfirst}0.399 & 23.26 & 5.37 \\ \bottomrule
\end{tabular}%
}
\caption{The completion performance on the SemanticKITTI dataset. Colors denote the \colorbox[RGB]{\colorfirst}{1st}, \colorbox[RGB]{\colorsecond}{2nd}, and \colorbox[RGB]{\colorthird}{3rd} best-performing model. The sampling time is estimated based on the official code and the provided checkpoints.}
\label{tab:completion1}
\end{table}

\begin{table}[t]
\centering
\resizebox{\columnwidth}{!}{%
\begin{tabular}{lrrrr}
\toprule
Model & \multicolumn{1}{c}{CD $\downarrow$} & \multicolumn{1}{c}{JSD $\downarrow$} & \multicolumn{1}{c}{EMD $\downarrow$} & \multicolumn{1}{c}{Times (s) $\downarrow$} \\ \midrule
LMSCNet~\cite{lmscnet} & 0.979 & 0.496 & - & 0.38 \\
LODE~\cite{lode} & 1.565 & 0.483 & - & 1.64 \\
MID~\cite{MID} & 0.637 & 0.476 & - & 6.23 \\
LiDiff~\cite{LiDiff} & 0.564 & 0.459 & \cellcolor[RGB]{\colorfirst}21.98  & \cellcolor[RGB]{\colorthird}29.18 \\
LiDiff (Refined)~\cite{LiDiff} & \cellcolor[RGB]{\colorthird}0.517 & \cellcolor[RGB]{\colorthird}0.446 & \cellcolor[RGB]{\colorthird}22.96 & 29.43 \\ \midrule
ScoreLiDAR & \cellcolor[RGB]{\colorsecond}0.472 & \cellcolor[RGB]{\colorsecond}0.444 & \cellcolor[RGB]{\colorsecond}22.80 & \cellcolor[RGB]{\colorfirst}4.98 \\
ScoreLiDAR (Refined) & \cellcolor[RGB]{\colorfirst}0.452 & \cellcolor[RGB]{\colorfirst}0.437 & 23.02 & \cellcolor[RGB]{\colorsecond}5.14 \\ \bottomrule
\end{tabular}%
}
\caption{The completion performance on the KITTI-360 dataset.
The meaning of notations is the same as those in \cref{tab:completion1}.
} 
\label{tab:completion2}
\end{table}

\begin{table}[t]
\centering
\resizebox{\columnwidth}{!}{%
\begin{tabular}{lrrrrrr}
\toprule
\multirow{2}{*}{Model} & \multicolumn{3}{c}{SemanticKITTI} & \multicolumn{3}{c}{KITTI360} \\ \cline{2-7} 
 & CD $\downarrow$ & JSD $\downarrow$ & EMD $\downarrow$ & CD $\downarrow$ & JSD $\downarrow$ & EMD $\downarrow$ \\ \midrule
ScoreLiDAR (Refined) & \cellcolor[RGB]{\colorfirst}0.342 & \cellcolor[RGB]{\colorfirst}0.399 & \cellcolor[RGB]{\colorfirst}23.26 & \cellcolor[RGB]{\colorfirst}0.452 & \cellcolor[RGB]{\colorfirst}0.437 & \cellcolor[RGB]{\colorfirst}23.02 \\
w/o Structural Loss (Refined) & 0.419 & 0.430 & 24.61 & 0.549 & 0.445 & 24.56 \\ \bottomrule
\end{tabular}%
}
\caption{Ablation study of the structural loss.}
\label{tab:ablation1}
\end{table}

\begin{table}[!t]
\centering
\resizebox{\columnwidth}{!}{%
\begin{tabular}{lrrrr}
\toprule
Model & \multicolumn{1}{c}{CD $\downarrow$} & \multicolumn{1}{c}{JSD $\downarrow$} & \multicolumn{1}{c}{EMD $\downarrow$} & \multicolumn{1}{c}{Time (s) $\downarrow$} \\ \midrule
LiDiff (50 steps)~\cite{LiDiff} & 0.434 & 0.444 & \cellcolor[RGB]{\colorfirst}22.15 & 30.38 \\
LiDiff (50 steps Refined)~\cite{LiDiff} & \cellcolor[RGB]{\colorthird}0.375 & 0.416 & \cellcolor[RGB]{\colorthird}23.16 & 30.55 \\
LiDiff (8 steps)~\cite{LiDiff} & 0.447 & 0.432 & 24.90 & 5.69 \\
LiDiff (8 steps Refined)~\cite{LiDiff} & 0.411 & 0.406 & 25.74 & 5.92 \\ \midrule
ScoreLiDAR (8 Steps Refined) & \cellcolor[RGB]{\colorsecond}0.342 & 0.399 & \cellcolor[RGB]{\colorsecond}23.14 & 5.37 \\
ScoreLiDAR (4 Steps Refined) & \cellcolor[RGB]{\colorfirst}0.326 & \cellcolor[RGB]{\colorthird}0.386 & 23.98 & \cellcolor[RGB]{\colorthird}3.23 \\
ScoreLiDAR (2 Steps Refined) & 0.403 & \cellcolor[RGB]{\colorfirst}0.379 & - & \cellcolor[RGB]{\colorsecond}1.85 \\
ScoreLiDAR (1 Steps Refined) & 0.548 & \cellcolor[RGB]{\colorsecond}0.384 & - & \cellcolor[RGB]{\colorfirst}1.10 \\ \bottomrule
\end{tabular}%
}
\caption{Ablation study of different sampling steps on the SemanticKITTI dataset.}
\label{tab:ablation2}
\end{table}


\section{Experiment}
\label{sec:experi}
In this part, we conduct a series of experiments to evaluate the effectiveness of the proposed ScoreLiDAR. We compare ScoreLiDAR with advanced models including LMSCNet~\cite{lmscnet}, LODE~\cite{lode}, MID~\cite{MID}, PVD~\cite{pvd} and LiDiff~\cite{LiDiff}. We first evaluate the performance of ScoreLiDAR in scene completion tasks (\cref{subsec:completion}). Secondly, we present the results of ablation studies showing the effectiveness of the structural loss and the performances of ScoreLiDAR given different sampling steps (\cref{subsec:ablation}). Finally, we further evaluate ScoreLiDAR with the qualitative analysis (\cref{subsec:qualitative}).

\begin{figure*}[t]
    \centering
    \includegraphics[width=0.95\linewidth]{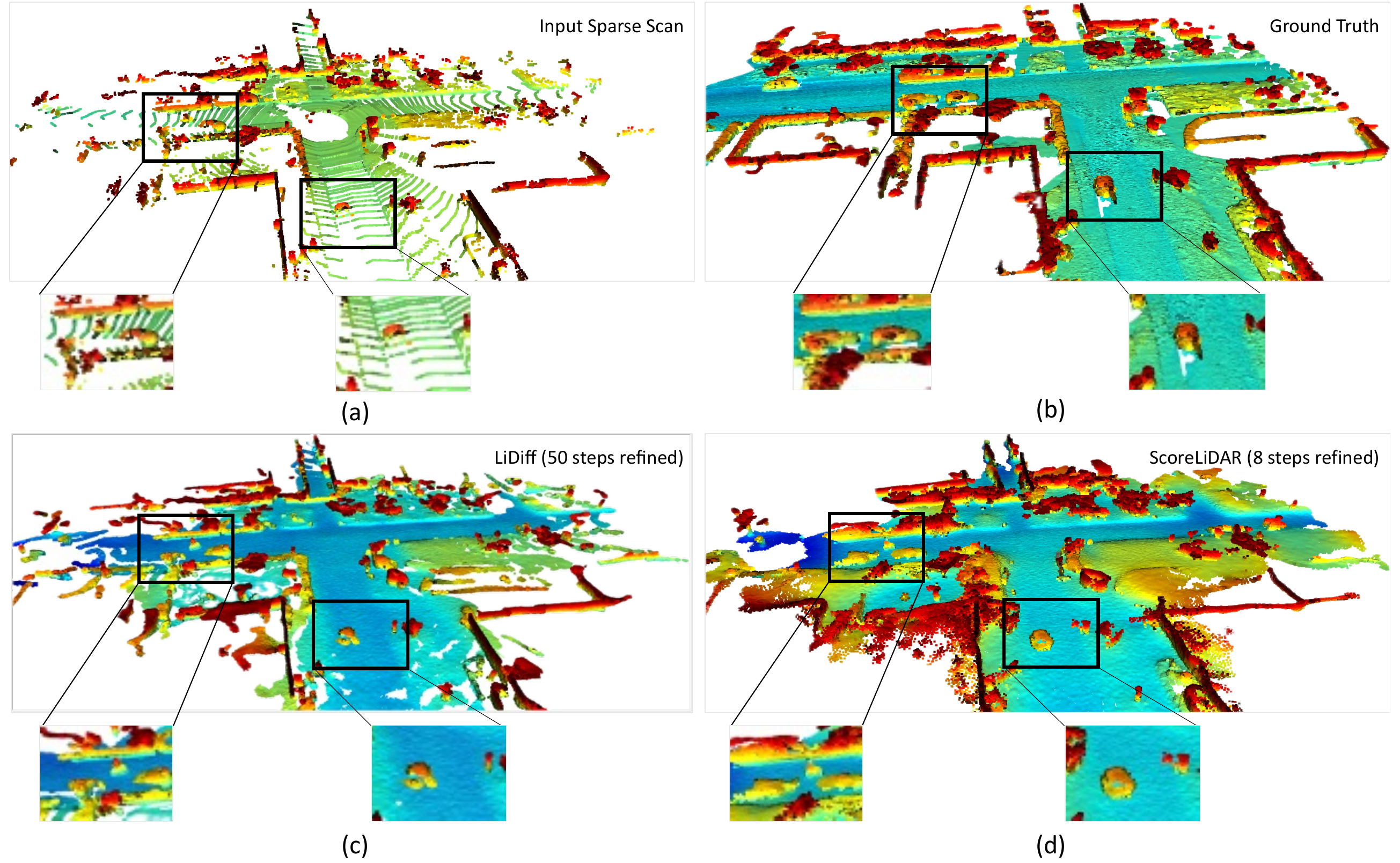}
    \caption{Qualitative results on KITTI-360. ScoreLiDAR achieves better completion than LiDiff~\cite{LiDMs} with fewer sampling steps.}
    \vspace{-1em}
    \label{fig:qualitative}
\end{figure*}

\subsection{Scene completion}
\label{subsec:completion}
We validate ScoreLiDAR on SemanticKITTI~\cite{semantickitti} and KITTI-360~\cite{kitti360} datasets. The existing SOTA LiDAR scene completion model LiDiff~\cite{LiDiff} is chosen as the teacher model. The student model shares the network architecture with the teacher model and is initialized by the teacher model. Moreover, we also use the refinement network in LiDiff~\cite{LiDiff} to refine the completed scene generated by the student model. We calculate the Chamfer Distance (CD)~\cite{CD}, the Jensen-Shannon Divergence (JSD)~\cite{JSD} and the Earth Mover's Distance (EMD)~\cite{EMD} to evaluate the similarity between the completed scene and the ground truth. The smaller the value of these metrics, the closer the completed scene is to the ground truth.

\cref{tab:completion1} shows that ScoreLiDAR achieves the optimal performance compared to the existing models in most cases on the SemanticeKITTI dataset. Compared to the SOTA method LiDiff~\cite{LiDiff} with refinement, which takes 30.55 seconds to complete a scene, ScoreLiDAR completes a scene in just 5.47 seconds (fivefold speedup) yet with $8\%$ improvement in CD,  $4\%$ in JSD, and comparable results in EMD.
The reason for the increase in EMD after refinement is the mismatch in the points numbers between $\mathcal{G}^0$ and $\mathcal{G}$, leading to a higher matching cost. The refinement process optimizes local details but alters the global point cloud distribution, extending the overall transfer path. Although LMSCNet~\cite{lmscnet} and LODE~\cite{lode} have faster completion speeds, their completion quality is significantly lower. The performance of ScoreLiDAR outperforms the teacher model LiDiff~\cite{LiDiff}. This is because ScoreLiDAR introduces a structural loss with scene-wise term and point-wise term, enabling the student model to effectively capture geometric structure information within LiDAR point cloud data during training. 
Results on KITTI-360 are shown in~\cref{tab:completion2}. ScoreLiDAR also achieves optimal performance in most cases and boasts a fivefold speedup with $12\%$ improvement in CD and $2\%$ in JSD compared to LiDiff~\cite{LiDiff}. 

\subsection{Ablation study}
\label{subsec:ablation}

In this part, we conduct the ablation study to verify the effectiveness of the structural loss in the training of the proposed ScoreLiDAR. We compared the scene completion performances of the proposed ScoreLiDAR with a variant that does not incorporate structural loss. The results are shown in~\cref{tab:ablation1}. The results show that the variant without structural loss exhibits lower performance in scene completion on both datasets. However, after considering the structural loss, the performance of ScoreLiDAR improves significantly, which achieves better performance on all metrics. This supports our discussion in~\cref{sec:method}, incorporating structural loss enables the student model to capture the geometric structure feature of 3D point clouds, thereby facilitating the effective distillation of the student model.


Furthermore, we compared the scene completion performance of ScoreLiDAR with different sampling steps, and the results are shown in~\cref{tab:ablation2}. It can be observed that as the sampling steps decrease from 8 to 1, the time required for ScoreLiDAR to complete a scene also decreases, with single-step sampling allowing a scene to be completed in only 1.1 seconds. With 8-step and 4-step sampling, ScoreLiDAR performs better than LiDiff. All metrics decay at 2-step and 1-step sampling, but in JSD ours still performs better than LiDiff. In summary, although the quality of scene completion decreases as the sampling steps are reduced, it still maintains performance comparable to or better than the existing model, achieving better performance and speed trade-off as in \cref{fig:CD_time}.

In addition, we compared the ablation results about the terms of structural loss, different keypoint selection methods, varying numbers of keypoints, and different values of $\lambda_{scene}$ and $\lambda_{point}$ and so on, which are presented in Sec.~S4.

\subsection{Qualitative analysis}
\label{subsec:qualitative}
\cref{fig:qualitative} shows the completed scenes by our proposed ScoreLiDAR and LiDiff~\cite{LiDiff} on KITTI-360. ScoreLiDAR achieves completion results with higher quality and greater fidelity in less time compared to LiDiff. For example, as shown in~\cref{fig:qualitative}(c), in the left region, there should be two vehicles as in~\cref{fig:qualitative}(b), but the scene completed by LiDiff only contains one. In contrast, the scene completed by ScoreLiDAR with only 8 steps in~\cref{fig:qualitative}(d) completes two vehicles and has clearer and more complete vehicle structures, making it closer to the ground truth.
\begin{figure*}
    \centering
    \includegraphics[width=0.95\linewidth]{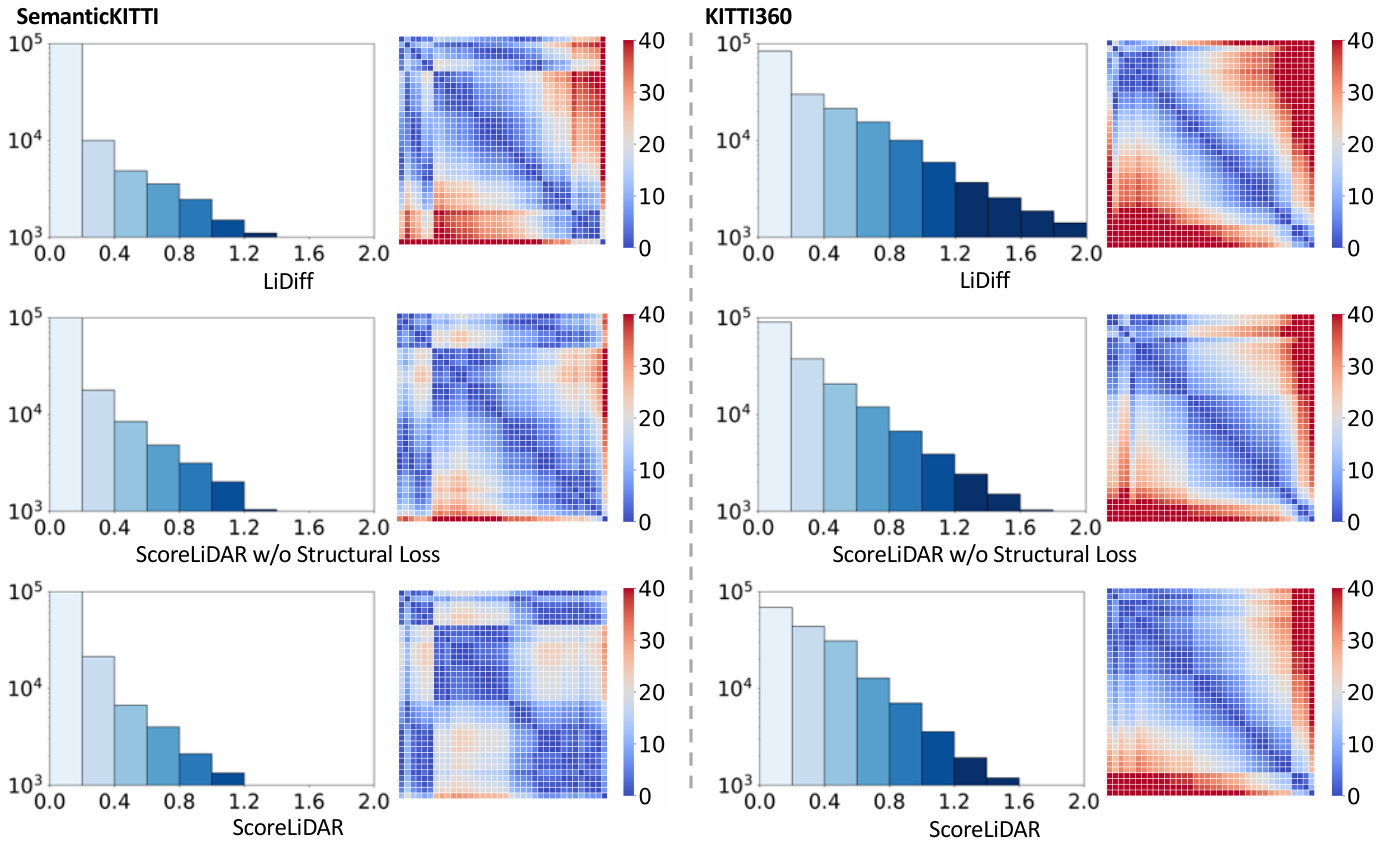}
    \caption{The qualitative analysis of structural loss. The bar chart shows the distribution of distances between corresponding points in the completed and ground truth scenes. A higher number of points with smaller distances demonstrates that the completed scene is closer to the ground truth. The heatmap represents the difference in distance matrices between the completed scene and the ground truth scene. Smaller values on the heatmap indicate that the completed scene is closer to the ground truth.}
    \vspace{-1em}
    \label{fig:heat}
\end{figure*}

To further demonstrate the effectiveness of ScoreLiDAR and the structural loss, we calculate the distance between the points in the completed scene and their corresponding points in the ground truth to evaluate the overall difference. We display the calculated results in bar chat in~\cref{fig:heat}. ScoreLiDAR has the highest number of points with smaller distances to their corresponding points in the ground truth. The results show that the scenes completed by ScoreLiDAR are closer to the ground truth overall, demonstrating higher fidelity. Moreover, we selected 36 corresponding key points from the ground truth and the completed scene using the method described in~\cref{subsec:structural_loss} and calculated the point distance matrices  $\mathcal{D}$ and $\mathcal{D}_G$. We then visualized the difference between $\mathcal{D}$ and $\mathcal{D}_G$ as a heatmap. As shown in~\cref{fig:heat}, on both datasets, the difference of point distance matrix between the completed scene of LiDiff~\cite{LiDiff} and the ground truth is the largest, followed by the ScoreLiDAR variant without the structural loss and the smallest difference is achieved by ScoreLiDAR. This also indicates that the scene completed by ScoreLiDAR is closer to the ground truth.

We also conduct a user study to evaluate the performance of ScoreLiDAR. We used ScoreLiDAR and LiDiff~\cite{LiDiff} to complete scenes based on the same input scans and asked users to choose the scene they believed was closer to the ground truth. ScoreLiDAR received a $65\%$ user preference over LiDiff~\cite{LiDiff}. This indicates that the detail and fidelity of the scenes completed by ScoreLiDAR more closely resemble the ground truth for most users. The details of the user study are shown in SM.



\section{Conclusion}
\label{sec:con}

\paragraph{Summary.} This paper proposes ScoreLiDAR, a novel distillation method tailored for 3D LiDAR scene completion diffusion models based on a bidirectional gradient guidance mechanism. By introducing the structural loss with scene-wise term and point-wise term, ScoreLiDAR trains the student model to effectively capture the holistic structure and the relative configuration of key points and achieve efficient and high-quality scene completion. 

\paragraph{Limitations.} While ScoreLiDAR achieves efficient, high-quality LiDAR scene completion, its performance is constrained by the teacher model. As the performance of the teacher model improves, so does the capability of the student model. Although we conducted a preliminary experiment on the semantic scene completion in the Supplementary Material, a more comprehensive exploration is still required.
Thus, further exploration is required to find a more effective method to improve the training process of ScoreLiDAR and avoid the limitations of the teacher model, achieving more efficient semantic LiDAR scene completion.

\section*{Acknowledgement}
This paper is supported by Provincial Key Research and Development Plan of Zhejiang Province under No. 2024C01250(SD2), National Natural Science Foundation of China (Grant No. 62006208) and Dream Set Off - Kunpeng $\&$ Ascend Seed Program.

{
    \small
    \bibliographystyle{ieeenat_fullname}
    \bibliography{main}
}

\clearpage
\appendix
\section{Experiment protocol}
\subsection{Dataset setup}
SemanticKITTI~\cite{semantickitti} dataset is a large-scale benchmark for 3D semantic segmentation in autonomous driving, extending the KITTI Odometry dataset with dense semantic annotations for over 43,000 LiDAR scans. It provides labels for 25 classes, such as “car,” “road,” and “building,” capturing diverse urban and rural scenes. The SemanticKITTI dataset consists of 22 sequences, where sequences 00-10 are densely annotated for each scan, enabling tasks such as semantic segmentation and semantic scene completion using sequential scans. Sequences 11-21 serve as the test set, showcasing diverse and challenging traffic situations and environment types to evaluate model performance in real-world autonomous driving scenarios. SemanticKITTI is widely used in research and serves as a critical resource for advancing LiDAR-based perception systems.

KITTI-360~\cite{kitti360} dataset is a comprehensive benchmark for 3D scene understanding in autonomous driving, capturing 360-degree panoramic imagery and 3D point clouds across diverse urban environments. It includes over 73 km of driving data with dense semantic annotations for both 2D (images) and 3D (point clouds), covering categories like “vehicles,” “buildings,” and “vegetation.” KITTI-360 provides high-resolution sensor data, including LiDAR, GPS/IMU, and stereo camera recordings, making it ideal for tasks such as 3D semantic segmentation, panoptic segmentation, and mapping in real-world driving scenarios.


\subsection{Evaluation metrics}
\paragraph{Chamfer Distance (CD)}~\cite{CD} is a metric used to measure the similarity between two sets of points, often employed for evaluating the quality of generated point clouds or geometric shapes. For two point sets $P$ and $Q$, the Chamfer Distance is defined as:
\begin{equation}
C D(P, Q)=\frac{1}{|P|} \sum_{p \in P} \min _{q \in Q}\|p-q\|^{2}+\frac{1}{|Q|} \sum_{q \in Q} \min _{p \in P}\|q-p\|^{2}
\label{eq:cd}
\end{equation}
The first term calculates the average squared distance from each point in $P$ to its nearest neighbour in $Q$. The second term calculates the average squared distance from each point in $Q$ to its nearest neighbour in $P$. Chamfer Distance evaluates how well two point sets approximate each other by considering their nearest neighbour distances in both directions. CD effectively captures local geometric features and exhibits strong robustness in local shape matching, which is commonly used in evaluating the matching and reconstruction of 3D point clouds.

\paragraph{Jensen-Shannon Divergence (JSD)}~\cite{JSD} is a symmetric measure of similarity between two probability distributions. It is a variation of the Kullback-Leibler (KL) divergence and is widely used in information theory, statistics, and machine learning. Given two probability distributions $P$ and $Q$ over the same domain, JSD is defined as:
\begin{equation}
    JSD(P \| Q)=\frac{1}{2} K L(P \| M)+\frac{1}{2} K L(Q \| M)
\end{equation}
Here $M = \frac{1}{2}(P + Q)$ is the average distribution, and $KL(P || M)$ is the Kullback-Leibler divergence. 

JSD measures how much $P$ and $Q$ diverge from their average distribution $M$. It is symmetric $(JSD(P || Q) = JSD(Q || P))$ and always produces a finite value in the range [0, 1] when using base-2 logarithms. Unlike KL divergence, JSD avoids issues with undefined values when probabilities are zero in one of the distributions. JSD is an efficient metric to evaluate the similarity between two distributions. The calculation of JSD in this paper is followed by Xiong~\textit{et al.}~\cite{JSD2}.

\subsection{Implementation details}
\label{sec:setup}
We choose the pre-trained LiDiff~\cite{LiDiff} model as the teacher model $\boldsymbol{\epsilon_\theta}$, the student model $G_{stu}$ and the auxiliary diffusion model $\boldsymbol{\epsilon}_\phi$ shares the same network architecture as the teacher model and are initialized by the teacher model. The ScoreLiDAR is trained on SemanticKITTI dataset.
The pre-trained diffusion model is provided by the official release of LiDiff~\cite{LiDiff}. For fair comparison, we follow LiDiff~\cite{LiDiff}'s training strategy. ScoreLiDAR is trained on sequences 00–07 and 09–10 of SemanticKITTI (does not train on KITTI-360), and evaluated on sequence 08 of SemanticKITTI and sequence 00 of KITTI-360.

For optimization, we use the Stochastic Gradient Descent (SGD) optimizer with the default parameters. The learning rate is set to $3e-5$ and the batch size is set to $1$. The training ratio between the student model and the auxiliary diffusion model is maintained at $1:1$. To reduce computational costs, when calculating the point-wise loss, we first randomly select $\frac{1}{10}$ of the points from the ground truth scene. Then, following the proposed method, we select the top $\frac{1}{3}$ points with the highest curvature from these points as the key points to calculate the distance matrix. That is, the final number of key points is $\frac{1}{30}$ of the total number of points in the ground truth scene. When calculating the $K$-nearest neighbours, we set $K = 180$. The weights of scene-wise loss $\lambda_{scene}$ and the point-wise loss $\lambda_{point}$ are set to $0.5$ and $0.01$, respectively. ScoreLiDAR requires only 50 iterations to achieve convergence, taking approximately 10 minutes on a single A40 GPU, which is highly efficient. Our model and code are publicly available on \url{https://github.com/happyw1nd/ScoreLiDAR}.


\section{Discussion}
\subsection{The effectiveness of the distillation on improving the completion efficiency}
In this part, we provide a detailed discussion about the efficiency of the proposed distillation method.

\paragraph{Why is it reasonable to initialize the student model and auxiliary diffusion model using the teacher model?} Firstly, such an initialization method is commonly used in existing methods~\cite{Diff-Instruct,zhang2024distribution,DMD,DMD2,consistency_model}. Secondly, the pre-trained teacher model $\boldsymbol{\epsilon}_\theta$ contains the information about the training distribution, initializing the student model with the teacher model to perform distillation is essentially a fine-tuning process for the teacher model, which can accelerate the efficiency of the distillation. Third, although the student model has the same parameters and the network structure as the teacher model, its sampling distribution is different from that of the teacher model due to the different sampling steps. The teacher model $\boldsymbol{\epsilon}_\theta$ (LiDiff~\cite{LiDiff} in this case) conducts the multi-step sampling. 
\begin{equation}
\label{eq:SM_reverse_process_multi}
    \mathcal{G}^{t-1}=\frac{1}{\sqrt{\alpha^{t}}}\left(\mathcal{G}^{t}-\frac{1-\alpha^{t}}{\sqrt{1-\bar{\alpha}^{t}}} \boldsymbol{\epsilon}_{\theta}\left(\mathcal{G}^{t}, \mathcal{P}, t\right)\right) + \sigma^t \boldsymbol{z}
\end{equation}
The teacher model $\boldsymbol{\epsilon}_\theta$ from LiDiff~\cite{LiDiff} conducts 50-steps sampling by repeating~\cref{eq:SM_reverse_process_multi}. The student model $G_{stu}$ conducts the single-step sampling. After $G_{stu}$ predicts the noise, a single-step denoising in~\cref{eq:SM_reverse_process} is performed to directly obtain the completed scene $\mathcal{G}^0$.
\begin{equation}
\label{eq:SM_reverse_process}
    \mathcal{G}^{0}=\frac{1}{\sqrt{\alpha^{t}}}\left(\mathcal{G}^{t}-\frac{1-\alpha^{t}}{\sqrt{1-\bar{\alpha}^{t}}} \boldsymbol{\epsilon}_{\theta}\left(\mathcal{G}^{t}, \mathcal{P}, t\right)\right)
\end{equation}
Thus, the single-step sampling scene of the student model is different from the multi-step sampling scene of the teacher model.

\paragraph{Why is the distillation loss effective? Why does the student model get optimized?} Firstly, our distillation loss is different from the standard loss of the diffusion models such as DDPM~\cite{DDPM}. The loss of DDPM is to directly predict the noise added to the training sample. Differently, in proposed ScoreLiDAR, the distillation loss utilizes the noise predicted by the student model $G_{stu}$ to perform a one-step sampling, resulting in a completed scene $\mathcal{G}^0$ different from the multi-step sampling of the teacher model. The completed scene $\mathcal{G}^0$ is then perturbed noise on a random timestep $t$ to obtain the noisy scene $\mathcal{G}^t$, and the difference between two score functions according to $\mathcal{G}^t$ is calculated to serve as the gradient for optimizing the student model, as shown in~\cref{eq:SM_G_gradient}
\begin{equation}
\label{eq:SM_G_gradient}
    \mathcal{L}_{KL} \approx \mathbb{E}_{t, \epsilon} \left[\|\boldsymbol{\epsilon}_{\theta}\left(\mathcal{G}^{t}, \mathcal{P}, t\right)- \boldsymbol{\epsilon}_{\phi}\left(\mathcal{G}^{t}, \mathcal{P}, t\right)\|^2_2\right]
\end{equation}

Secondly, although the auxiliary diffusion model $\boldsymbol{\epsilon}_\phi$ is initilized from the teacher model $\boldsymbol{\epsilon}_\theta$, the gradient in~\cref{eq:SM_G_gradient} is non-zero and efficient. Recall that the auxiliary diffusion model$\boldsymbol{\epsilon}_\phi$  and the student model $G_{stu}$ are trained alternately. The auxiliary diffusion model $\boldsymbol{\epsilon}_\phi$ is first trained on $\mathcal{G}^0$ to fit the one-step sampling distribution, which is different from the pre-trained distribution of the teacher model. Although the auxiliary diffusion model $\boldsymbol{\epsilon}_\phi$ is also initilized from the teacher model, its parameter will be updated and become different from the teacher model's after one optimization. Then, when $G_{stu}$ is optimized using the gradient in~\cref{eq:SM_G_gradient}, the output of $\boldsymbol{\epsilon}_{\theta}\left(\mathcal{G}^{t}, \mathcal{P}, t\right)$ is naturally different from the output of $\boldsymbol{\epsilon}_{\phi}\left(\mathcal{G}^{t}, \mathcal{P}, t\right)$ due to the optimization of $\boldsymbol{\epsilon}_{\phi}$. Thus, as mentioned in Sec.~4.1 of the main paper, the non-zero gradient in~\cref{eq:SM_G_gradient} will optimize the distribution of $G_{stu}$ moving towards the pre-trained distribution of the teacher model. Then, $\boldsymbol{\epsilon}_{\phi}$ and $G_{stu}$ are optimized in turn to convergence.
%

\paragraph{Why is one-step sampling used during training, while few-step sampling is used during inference?} Firstly, in traditional diffusion models like DDPM~\cite{DDPM} and DDIM~\cite{DDIM}, they directly predict the whole noise added in the diffusion process during the training, whereas a multi-step sampling process is performed during sampling. Meanwhile, different sampling methods allow for sampling with different numbers of steps. Intuitively, this indicates that although the training objective remains the same, the sampling can be performed with different numbers of steps. A more profound explanation is from solving the stochastic differential equation as in the diffusion model. Moreover, Consistency Model~\cite{consistency_model} also has a similar setting; its student model performs one-step generation during training but conducts multi-step generation during sampling. We adopted a similar principle as in the Consistency Model. 
The one-step sampling in our training is to increase the fidelity of the resulting sample given different noisy samples, while in inference multi-step sampling would gradually refine the final result. 
Therefore, although the student model performs one-step sampling during training, the quality of scene completion can be improved by increasing the number of sampling steps during the inference. 
More visually, the one-step generation procedure of the Consistency Model is ``noise$\rightarrow$image'', and the multi-step generation procedure is ``noise$\rightarrow$noisy image$\rightarrow$$\cdots$$\rightarrow$image''. In this paper, the one-step sampling of student model $G_{stu}$ during training is ``noise$\rightarrow$predicted noise$\rightarrow$completed scene'' and the few-step sampling during inference is ``noise$\rightarrow$predicted noise$\rightarrow$noisy scene$\rightarrow$predicted noise$\rightarrow$noisy scene$\rightarrow$predicted noise$\rightarrow$$\cdots$$\rightarrow$completed scene''. 
In summary, it is reasonable to use one-step sampling during training and multi-step sampling during inference.

\subsection{The differents of scene-wise loss and Chamfer Distance}
The scene-wise loss has the following form
\begin{equation}
    \label{eq:SM_scene_wise}
    \mathcal{L}_{scene} = \frac{1}{|\mathcal{G}^0|} \sum_{\boldsymbol{p}^0_i \in \mathcal{G}^0} \min _{\boldsymbol{p} \in \mathcal{G}}\|\boldsymbol{p}^0_i-\boldsymbol{p}\|^{2}
\end{equation}
The scene-wise loss is part of the Chamfer Distance. The reason for using only part of the Chamfer Distance (CD) is that the optimization objective of scene-wise loss is to ensure that the points in the generated scene $\mathcal{G}^0$ are as close as possible to the corresponding points in the ground truth. The unused term in CD matches each point in ground truth with its nearest point in generated scene $\mathcal{G}^0$, which may lead to points in $\mathcal{G}^0$ being pushed toward the average position of non-existent matching points in the ground truth. This effect is detrimental to scene completion.

\subsection{Discussion on the significance of this study}
Firstly, we discuss the significance of this study. For autonomous vehicles, accurately recognizing and perceiving their surrounding environment during operation is critical~\cite{voxformer,emergent}. This is particularly important for identifying objects that may affect the vehicle’s movement, such as other vehicles, pedestrians, traffic cones, and signposts~\cite{jung20183d}. The accurate and efficient recognition of these objects is essential for the safe operation of autonomous vehicles. However, the scan data obtained by onboard LiDAR is sparse~\cite{MID,lode}, and it is difficult to identify key objects such as vehicles from the magnified regions of the sparse scan. Autonomous vehicles cannot obtain sufficient information about the driving environment from these sparse LiDAR scans~\cite{LiDMs,lee2023diffusion}. Therefore, it is necessary to use appropriate methods to complete the sparse LiDAR scans.

LiDiff~\cite{LiDiff} uses DDPM~\cite{DDPM} models to complete 3D LiDAR scenes, achieving impressive results. However, due to the inherent characteristics of diffusion models, LiDiff~\cite{LiDiff} requires approximately 30 seconds to complete a single scene, limiting its applicability in autonomous vehicles. In contrast, the proposed ScoreLiDAR can complete a scene in almost 5 seconds, more than 5 times faster than LiDiff~\cite{LiDiff}, while achieving higher completion quality. Thus, with the scenes completed by the proposed ScoreLiDAR, autonomous vehicles can more easily recognize critical objects in their driving environment, enabling safer and more effective navigation.

\section{Additional completed scenes}

\begin{figure*}
    \centering
    \includegraphics[width=\linewidth]{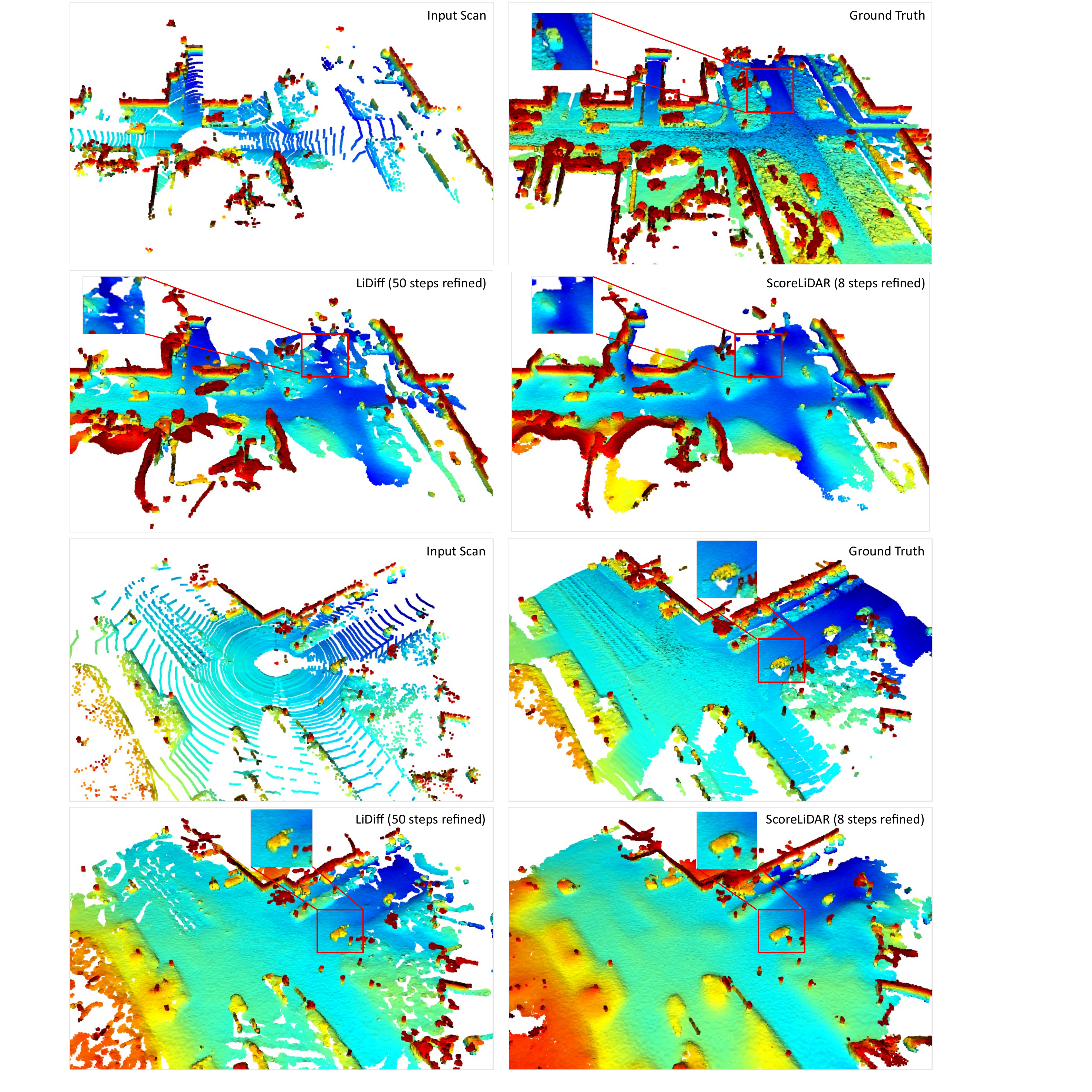}
    \caption{Completed samples of ScoreLiDAR from KITTI-360 dataset.}
    \label{fig:addition1}
\end{figure*}

\begin{figure*}
    \centering
    \includegraphics[width=\linewidth]{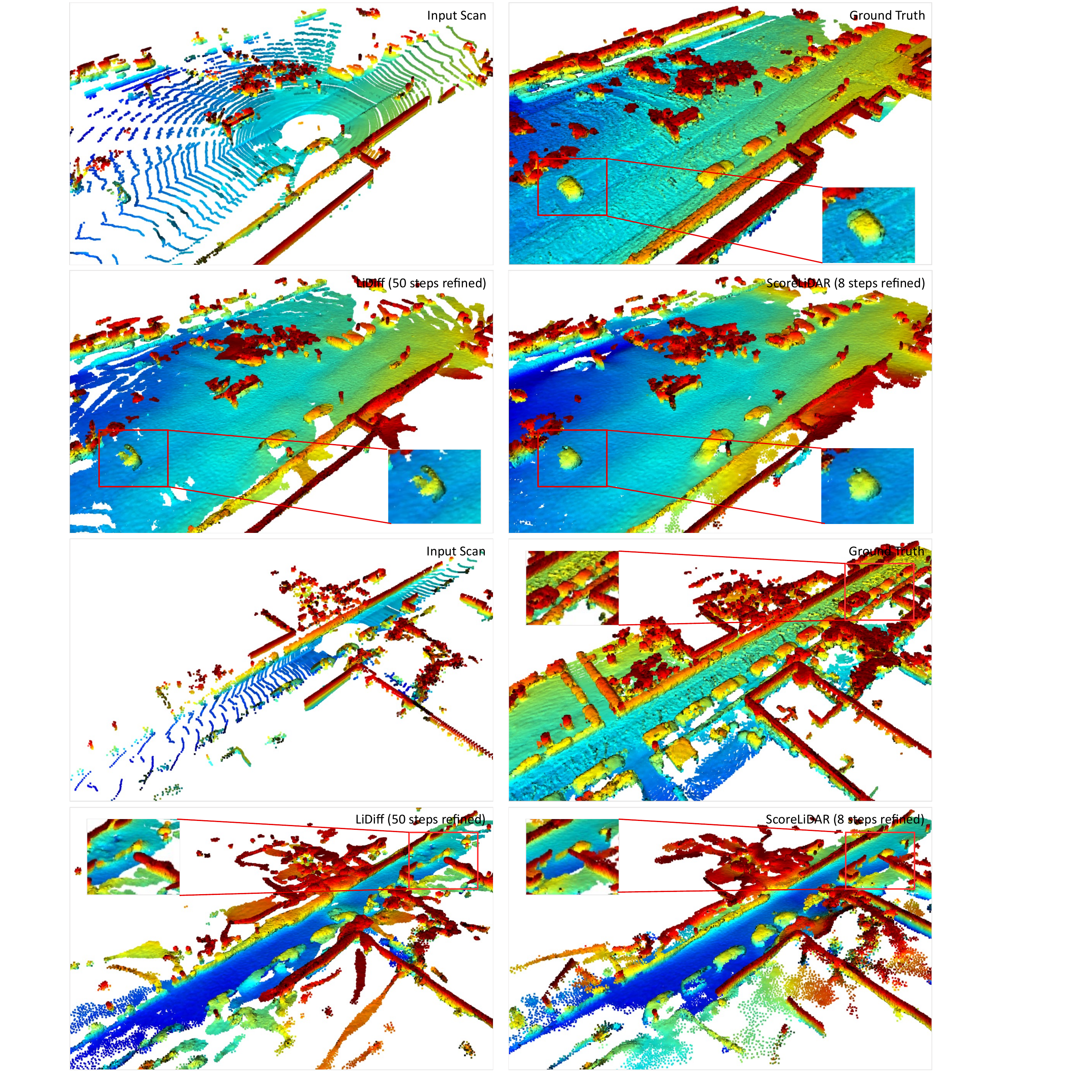}
    \caption{Completed samples of ScoreLiDAR from SemanticKITTI dataset.}
    \label{fig:addition2}
\end{figure*}

\cref{fig:addition1} and \cref{fig:addition2} show additional completed scenes by the proposed ScoreLiDAR and compare them with the scenes completed by LiDiff~\cite{LiDiff}.
\section{Additional experiment results}

\subsection{More ablation study of structural loss}
\begin{table}[t]
\centering
\resizebox{\columnwidth}{!}{%
\begin{tabular}{lrrrrrr}
\toprule
\multirow{2}{*}{Model} & \multicolumn{3}{c}{SemanticKITTI} & \multicolumn{3}{c}{KITTI360} \\ \cline{2-7} 
 & CD $\downarrow$ & JSD $\downarrow$ & EMD $\downarrow$ & CD $\downarrow$ & JSD $\downarrow$ & EMD $\downarrow$ \\ \midrule
ScoreLiDAR & \cellcolor[RGB]{\colorfirst}0.342 & \cellcolor[RGB]{\colorfirst}0.399 &  \cellcolor[RGB]{\colorfirst}23.26 & \cellcolor[RGB]{\colorfirst}0.452 & \cellcolor[RGB]{\colorfirst}0.437 & \cellcolor[RGB]{\colorfirst}23.02 \\
w/o Point-wise loss & \cellcolor[RGB]{\colorsecond}0.351 & \cellcolor[RGB]{\colorsecond}0.414 & \cellcolor[RGB]{\colorsecond}23.37 & \cellcolor[RGB]{\colorthird}0.485 & 0.486 & \cellcolor[RGB]{\colorsecond}23.29\\ 
w/o Scene-wise loss & \cellcolor[RGB]{\colorthird}0.367 & \cellcolor[RGB]{\colorthird}0.422 & 24.89 & \cellcolor[RGB]{\colorsecond}0.477 & \cellcolor[RGB]{\colorthird}0.451 & 24.69 \\ 
w/o Structural Loss & 0.419 & 0.430 & \cellcolor[RGB]{\colorthird}24.13 & 0.549 & \cellcolor[RGB]{\colorsecond}0.445 & \cellcolor[RGB]{\colorthird}24.56 \\ 
\bottomrule
\end{tabular}%
}
\caption{Ablation study of the scene-wise and point-wise loss. The metrics refer to the performance with refinement. Colors denote the \colorbox[RGB]{\colorfirst}{1st}, \colorbox[RGB]{\colorsecond}{2nd}, and \colorbox[RGB]{\colorthird}{3rd} best-performing model.}
\label{tab:SM_ablation1}
\end{table}

\begin{table}[t]
\centering
\resizebox{\columnwidth}{!}{%
\begin{tabular}{lrrrrrr}
\toprule
\multirow{2}{*}{ScoreLiDAR} & \multicolumn{3}{c}{SemanticKITTI} & \multicolumn{3}{c}{KITTI360} \\ \cline{2-7} 
 & CD $\downarrow$ & JSD $\downarrow$ & EMD $\downarrow$ & CD $\downarrow$ & JSD $\downarrow$ & EMD $\downarrow$ \\ \midrule
$\lambda_{scene}=0.5, \lambda_{point}=0.01$& \cellcolor[RGB]{\colorfirst}0.342 & \cellcolor[RGB]{\colorfirst}0.399 & \cellcolor[RGB]{\colorfirst}23.26 & \cellcolor[RGB]{\colorfirst}0.452 & \cellcolor[RGB]{\colorfirst}0.437 & \cellcolor[RGB]{\colorfirst}23.02\\
$\lambda_{scene}=0.05, \lambda_{point}=0.01$ & \cellcolor[RGB]{\colorsecond}0.354 & \cellcolor[RGB]{\colorsecond}0.409 & \cellcolor[RGB]{\colorthird}23.40 & \cellcolor[RGB]{\colorsecond}0.494 & \cellcolor[RGB]{\colorsecond}0.457 & \cellcolor[RGB]{\colorthird}23.21\\ 
$\lambda_{scene}=0.5, \lambda_{point}=0.1 $ & \cellcolor[RGB]{\colorthird}0.358 & \cellcolor[RGB]{\colorthird}0.419 & \cellcolor[RGB]{\colorsecond}23.27 & \cellcolor[RGB]{\colorthird}0.539 & \cellcolor[RGB]{\colorthird}0.474 & \cellcolor[RGB]{\colorsecond}23.09\\ 
\bottomrule
\end{tabular}%
}
\caption{Ablation study of the different weights of the scene-wise and point-wise loss. The first row refers to the default configuration of the ScoreLiDAR. The metrics refer to the performance with refinement.}
\label{tab:SM_ablation2}
\end{table}

\begin{table}[t]
\centering
\begin{tabular}{lcccc}
\toprule
\multirow{2}{*}{Model} & \multicolumn{2}{c}{SemanticKITTI} & \multicolumn{2}{c}{KITTI360} \\ \cline{2-5} 
 & CD $\downarrow$ & JSD $\downarrow$ & CD $\downarrow$ & JSD $\downarrow$ \\ \midrule
LiDiff (Refined) & \cellcolor[RGB]{\colorfirst}0.375 & \cellcolor[RGB]{\colorfirst}0.416 & \cellcolor[RGB]{\colorfirst}0.517 & \cellcolor[RGB]{\colorfirst}0.446 \\
w/ Structural loss & 0.399 & 0.426 & 0.535 & 0.450 \\ 
\bottomrule
\end{tabular}%
\caption{Ablation study of training LiDiff~\cite{LiDiff} with structural loss.}
\label{tab:SM_ablation3}
\end{table}

To further validate the effectiveness of the structural loss, we evaluated the performance of variants trained with only point-wise loss or scene-wise loss and compared them with default ScoreLiDAR. As shown in~\cref{tab:SM_ablation1}, compared to the default ScoreLiDAR, the performance of variants trained with only scene-wise loss or point-wise loss decreased. However, compared to the variants without structural loss, the variants using only one type of loss still showed improved completion performance. 
These results confirm the effectiveness of the structural loss in the distillation process.


Additionally, we investigated the impact of different weights of scene-wise and point-wise loss on the completion quality. The results are shown in~\cref{tab:SM_ablation2}. It can be observed that reducing $\lambda_{scene}$ or increasing $\lambda_{point}$ leads to a decline in the performance of ScoreLiDAR but still achieves a comparable performance. 
This verifies the effectiveness of the proposed structural loss in improving the completion performance of the student model.

Finally, we trained LiDiff using structural loss to investigate whether structural loss can enhance the performance of LiDiff. The results are shown in~\cref{tab:SM_ablation3}. Training LiDiff~\cite{LiDiff} with structural loss does not result in a performance improvement. This may be because structural loss is not suitable for direct addition to the training loss of LiDiff~\cite{LiDiff}, \textit{i.e.} the traditional diffusion model training loss. 

\subsection{Ablation study of different key point number}
\begin{table}[t]
\centering
\resizebox{\columnwidth}{!}{%
\begin{tabular}{lrrrrrr}
\toprule
\multirow{2}{*}{ScoreLiDAR} & \multicolumn{3}{c}{SemanticKITTI} & \multicolumn{3}{c}{KITTI360} \\ \cline{2-7} 
 & CD $\downarrow$ & JSD $\downarrow$ & EMD $\downarrow$ & CD $\downarrow$ & JSD $\downarrow$ & EMD $\downarrow$\\ \midrule
$n=1/20$ & \cellcolor[RGB]{\colorfirst}0.329 & \cellcolor[RGB]{\colorfirst}0.392 & \cellcolor[RGB]{\colorsecond}24.35 & 0.528 & \cellcolor[RGB]{\colorthird}0.475 & \cellcolor[RGB]{\colorsecond}24.01\\ 
$n=1/30$ & \cellcolor[RGB]{\colorsecond}0.342 & \cellcolor[RGB]{\colorsecond}0.399 & \cellcolor[RGB]{\colorfirst}23.26 & \cellcolor[RGB]{\colorfirst}0.452 & \cellcolor[RGB]{\colorfirst}0.437 & \cellcolor[RGB]{\colorfirst}23.02 \\
$n=1/60$ & \cellcolor[RGB]{\colorthird}0.346 & \cellcolor[RGB]{\colorthird}0.409 & \cellcolor[RGB]{\colorthird}25.19 & \cellcolor[RGB]{\colorsecond}0.452 & \cellcolor[RGB]{\colorsecond}0.471& \cellcolor[RGB]{\colorthird}25.05\\  
$n=1/70$ & 0.428 & 0.454 & 25.60 & \cellcolor[RGB]{\colorthird}0.466 & 0.479 & 25.44 \\
\bottomrule
\end{tabular}%
}
\caption{Ablation study of different key points number. The result of $n=1/30$ refers to the default configuration of the ScoreLiDAR. The metrics refer to the performance with refinement.}
\label{tab:SM_ablation4}
\end{table}

As mentioned in~\cref{sec:setup}, the optimal number of the key point is set to the $\frac{1}{30}$ of the total number of points in the ground truth. To investigate the impact of different numbers of key points on the completion performance of ScoreLiDAR, we decreased the number of key points for model training and evaluated the completion performance. As shown in~\cref{tab:SM_ablation4}, the final performance of ScoreLiDAR is positively correlated with the number of key points. When the number of key points decreases, the performance of ScoreLiDAR declines. This is because an insufficient number of key points causes the point-wise loss to fail in effectively capturing the relative positional information between key points, preventing the student model from learning the local geometric structure, and thereby reducing the completion quality. However, when the number of key points is too large, it can easily cause an out-of-memory issue and reduce training efficiency. Therefore, this paper sets $n = \frac{1}{30}$.


\subsection{Ablation study of different key point selection method}
We compare different key point selection methods including random selection and farthest selection with the proposed selection method based on curvature. The results in~\cref{tab:selection} show that using the proposed selection method based on curvature achieves the optimal performance than other selection methods.
\begin{table}[t]
\centering
\begin{tabular}{lrrr}
\toprule
{Model} & \multicolumn{1}{c}{{CD $\downarrow$}} & \multicolumn{1}{c}{{JSD $\downarrow$}} & \multicolumn{1}{c}{{EMD $\downarrow$}} \\ \midrule
Random selection & \cellcolor[RGB]{\colorsecond}0.384 & \cellcolor[RGB]{\colorsecond}0.433 & \cellcolor[RGB]{\colorfirst}23.23 \\
farthest selection & \cellcolor[RGB]{\colorthird}0.442 & \cellcolor[RGB]{\colorthird}0.459 & \cellcolor[RGB]{\colorthird}23.63 \\ 
ScoreLiDAR & \cellcolor[RGB]{\colorfirst}0.342 & \cellcolor[RGB]{\colorfirst}0.399 & \cellcolor[RGB]{\colorsecond}23.26 \\ \bottomrule
\end{tabular}%
\caption{Comparison of different selection methods on SemanticKITTI dataset. The performance of the proposed selection method is better than others.}
\label{tab:selection}
\end{table}

\subsection{Ablation study of different sampling steps on KITTI-360}
We also conduct the ablation study of different sampling steps on the KITTI-360 dataset. The results are shown in~\cref{tab:ablation3}. Similar to the results on the SemanticKITTI dataset, as the number of sampling steps decreases, the time required for ScoreLiDAR to complete a scene is reduced. Although the completion performance declines slightly, it remains comparable to that of existing SOTA models.

\begin{table}[!t]
\centering
\resizebox{\columnwidth}{!}{%
\begin{tabular}{lrrrr}
\toprule
Model & \multicolumn{1}{c}{CD $\downarrow$} & \multicolumn{1}{c}{JSD $\downarrow$} & \multicolumn{1}{c}{EMD $\downarrow$} & \multicolumn{1}{c}{Time (s) $\downarrow$} \\ \midrule
LiDiff (50 steps)~\cite{LiDiff} & 0.564 & 0.549 & \cellcolor[RGB]{\colorfirst}21.98 & 29.18\\
LiDiff (50 steps Refined)~\cite{LiDiff} & \cellcolor[RGB]{\colorthird}0.517 & \cellcolor[RGB]{\colorsecond}0.446 & \cellcolor[RGB]{\colorsecond}22.96 & 29.43 \\
LiDiff (8 steps)~\cite{LiDiff} & 0.619 & 0.471 & 24.85 & 5.46 \\
LiDiff (8 steps Refined)~\cite{LiDiff} & 0.550 & 0.462 & 25.49 & 5.77 \\ \midrule
ScoreLiDAR (8 Steps) & \cellcolor[RGB]{\colorfirst}0.452 & \cellcolor[RGB]{\colorfirst}0.437 & \cellcolor[RGB]{\colorthird}23.02 & 5.14 \\
ScoreLiDAR (4 Steps) & \cellcolor[RGB]{\colorsecond}0.482 & 0.461 & 23.76 & \cellcolor[RGB]{\colorthird}3.16 \\
ScoreLiDAR (2 Steps) & 0.525 & \cellcolor[RGB]{\colorthird}0.457 & - & \cellcolor[RGB]{\colorsecond}1.69 \\
ScoreLiDAR (1 Steps) & 0.750 & 0.478 & - & \cellcolor[RGB]{\colorfirst}1.03 \\ \bottomrule
\end{tabular}%
}
\caption{Ablation study of different sampling steps on the KITTI-360 dataset. The
metrics of ScoreLiDAR refer to the performance with refinement.}
\label{tab:ablation3}
\end{table}

\subsection{User study}
The user study is conducted to verify the completion performance of ScoreLiDAR further. We first used ScoreLiDAR and the current SOTA method LiDiff~\cite{LiDiff} to complete the same 30 input LiDAR scans, resulting in 30 pairs of completed scenes. We then randomly recruited seven volunteers and guided each to evaluate the detail and fidelity of these 30 pairs of scene images, selecting the one they believed to be closer to the ground truth. The seven volunteers included five men and two women, aged 24–30, with five participants having research backgrounds related to autonomous driving or LiDAR scene completion and the remaining two participants having backgrounds related to artificial intelligence. They were given unlimited time for the evaluation, but the average completion time for all volunteers was 30 minutes.

\begin{table}[t]
\centering
\begin{tabular}{lr}
\toprule
{Model} & \multicolumn{1}{c}{{User preference $\uparrow$}}  \\ \midrule
LiDiff~\cite{LiDiff} &  35\% \\
ScoreLiDAR & 65\%  \\ \bottomrule
\end{tabular}%
\caption{Results of user study. Our ScoreLiDAR outperforms the existing SOTA model.}
\label{tab:user}
\end{table}

The result of the user study is shown in~\cref{tab:user}. 
Compared to LiDiff, ScoreLiDAR received a $65\%$ user preference, surpassing the majority threshold. This indicates that, in the eyes of most users, the detail and fidelity of the scenes completed by ScoreLiDAR more closely resemble the ground truth. The results of the user study further demonstrate the effectiveness of ScoreLiDAR in LiDAR scene completion.

\subsection{Visualization of key points}
\begin{figure*}
    \centering
    \includegraphics[width=\linewidth]{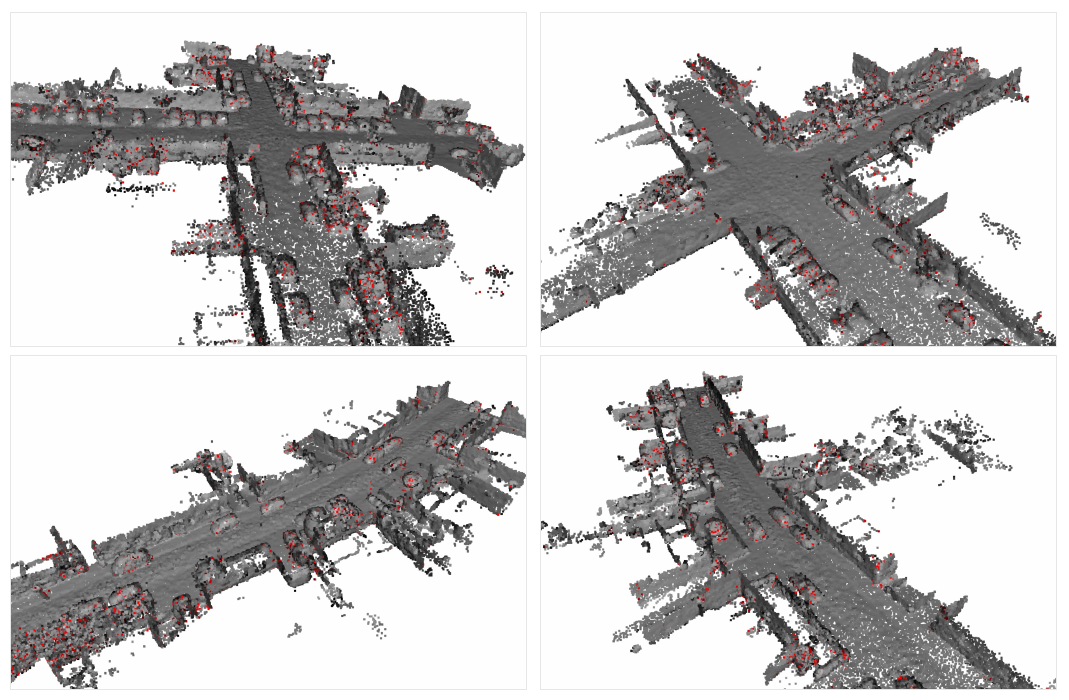}
    \caption{The visualization of the selected key points. Red points refer to the key points selected by the proposed method.}
    \label{fig:keypoints}
\end{figure*}

To validate the feasibility of our proposed key point selection method, we visualized the selected key points in the ground truth scene. As shown in~\cref{fig:keypoints}, the red key points are mostly distributed on walls, traffic cones, cars, and corners, while smooth areas such as the road surface have no key points. These key points are crucial for expressing the details of 3D LiDAR scenes. Selecting these points to compute the point-wise loss allows the student model to more easily capture the relative configuration information between key points, thereby better completing key objects in the scene.

\begin{table}[t]
\centering
\begin{tabular}{lrrr}
\toprule
\multirow{2}{*}{Model} & \multicolumn{3}{c}{SemanticKITTI (IoU) \% $\uparrow$} \\ \cline{2-4} 
 & \multicolumn{1}{c}{0.5m} & \multicolumn{1}{c}{0.2m} & \multicolumn{1}{c}{0.1m} \\ \midrule
LMSCNet~\cite{lmscnet} & 32.23 & 23.05 & 3.48 \\
LODE~\cite{lode} & \cellcolor[RGB]{\colorsecond}{43.56} & \cellcolor[RGB]{\colorfirst}{47.88} & 6.06 \\
MID~\cite{MID} & \cellcolor[RGB]{\colorfirst}{45.02} & \cellcolor[RGB]{\colorsecond}{41.01} & \cellcolor[RGB]{\colorsecond}{16.98} \\
PVD~\cite{pvd} & 21.20 & 7.96 & 1.44 \\
LiDiff~\cite{LiDiff} & \cellcolor[RGB]{\colorthird}{42.49} & 33.12 & 11.02 \\
LiDiff (Refined)~\cite{LiDiff} & 40.71 & \cellcolor[RGB]{\colorthird}{38.92} & \cellcolor[RGB]{\colorfirst}{24.75} \\ \midrule
ScoreLiDAR & 38.43 & 25.75 & 8.34 \\
ScoreLiDAR (Refined) & 37.33 & 29.57 & \cellcolor[RGB]{\colorthird}{15.63} \\ \bottomrule
\end{tabular}%
\caption{The IoU evaluation results on the SemanticKITTI dataset.}
\label{tab:occupancy1}
\end{table}

\begin{table}[tb]
\centering
\begin{tabular}{lrrr}
\toprule
\multirow{2}{*}{Model} & \multicolumn{3}{c}{KITTI-360 (IoU) \% $\uparrow$} \\ \cline{2-4} 
 & \multicolumn{1}{c}{0.5m} & \multicolumn{1}{c}{0.2m} & \multicolumn{1}{c}{0.1m} \\ \midrule
LMSCNet~\cite{lmscnet} & 25.46 & 16.35 & 2.99 \\
LODE~\cite{lode} & \cellcolor[RGB]{\colorthird}{42.08} & \cellcolor[RGB]{\colorfirst}{42.63} & 5.85 \\
MID~\cite{MID} & \cellcolor[RGB]{\colorfirst}{44.11} & \cellcolor[RGB]{\colorsecond}{36.38} & \cellcolor[RGB]{\colorthird}{15.84} \\
LiDiff~\cite{LiDiff} & \cellcolor[RGB]{\colorsecond}{42.22} & 32.25 & 10.80 \\
LiDiff (Refined)~\cite{LiDiff} & 40.82 & \cellcolor[RGB]{\colorthird}{36.08} & \cellcolor[RGB]{\colorfirst}{21.34} \\ \midrule
ScoreLiDAR & 36.82 & 25.49 & 9.70 \\
ScoreLiDAR (Refined) & 33.29 & 28.60 & \cellcolor[RGB]{\colorsecond}{15.95} \\ \bottomrule
\end{tabular}%
\caption{The IoU evaluation results on the KITTI-360 dataset.}
\label{tab:occupancy2}
\end{table}

\subsection{Experiments on semantic scene completion}
The objective of this paper is to propose a foundational method for distillation acceleration applicable to various LiDAR scene completion diffusion models. Semantic scene completion is not within the scope of the fundamental experiments considered in this paper. In spite of this, to validate the generalizability of the proposed method, we use SemCity~\cite{semcity} as the teacher model to verify the effectiveness of ScoreLiDAR on semantic scene completion tasks. Because the pre-trained models and code for the metric computation of SemCity are not publicly available, we retrained SemCity based on the official implementation and reproduced the metric computation ourselves. \cref{tab:ssc} shows the results. In semantic scene completion tasks, the proposed ScoreLiDAR still shows better completion quality than that of the teacher model.
\begin{table}[t]
\centering
\begin{tabular}{lrr}
\toprule
{Model} & \multicolumn{1}{c}{{FID $\downarrow$}} & \multicolumn{1}{c}{{KID $\downarrow$}} \\ \midrule
SemCity~\cite{semcity} & 88.52 & 0.11 \\
ScoreLiDAR & 81.76 & 0.09 \\ \bottomrule
\end{tabular}%
\caption{Results of semantic scene completion. Our ScoreLiDAR shows better performance.}
\label{tab:ssc}
\end{table}

\subsection{Experiments on scene occupancy}
We calculate the Intersection-Over-Union (IoU)~\cite{iou} to evaluate the occupancy of the completed scene compared with the ground truth scene. IoU represents the degree of overlap between the voxels in the completed scene and those in the ground truth scene. A higher IoU value indicates a higher completeness of the completed scene. During the evaluation, we considered three different voxel resolutions: $0.5m$, $0.2m$, and $0.1m$. The smaller the voxel resolution, the more fine-grained details are considered in the evaluation metrics, and vice versa. However, IoU is a voxel-based metric, and in some voxel-based LiDAR scene completion methods, it can serve as an accurate measure of completion quality. In contrast, ScoreLiDAR is a point cloud-based completion method, which differs from traditional voxel-based approaches. As a result, IoU may introduce bias when evaluating the completion quality. Therefore, here we provide IoU results solely as a relative reference.

\cref{tab:occupancy1} and \cref{tab:occupancy2} show the IoU of ScoreLiDAR and existing models. Under low voxel resolutions, ScoreLiDAR achieves comparable IoU values, meaning ScoreLiDAR generates dense and accurate point clouds. When the voxel resolutions become higher, the performance of ScoreLiDAR declines. As mentioned above, the existing method is mainly based on signed distance fields, which implement the scene completion using a voxel representation. ScoreLiDAR is point-level scene completion with the input of point clouds obtained from LiDAR scans, which works better at smaller voxel resolutions. 

\section{Theoretical demonstration}
The gradient of the student model is~\cref{eq:smgradient}. 
\begin{equation}
\label{eq:smgradient}
\begin{aligned}
    & \nabla_\eta D_{\mathrm{KL}} \left(p_G^{t} \left( \mathcal{G}^t \right) \| q^{t}\left(\mathcal{G}^t\right)\right) \\ & = 
    \mathbb{E}_{t,\epsilon} \left[ \nabla_{\mathcal{G}^t}\log p_G^{t}\left(\mathcal{G}^t\right) - \nabla_{\mathcal{G}^t} \log q^{t}\left(\mathcal{G}^t\right)\right] \frac{\partial \mathcal{G}^t}{\partial \eta}
\end{aligned}
\end{equation}

As proposed in ScoreSDE~\cite{ScoreSDE}, the log likelihood $\nabla_{\mathcal{G}^t} \log q^{t}\left(\mathcal{G}^t\right)$ can be approximated by the predicted noise $\hat{\epsilon}$ with $\nabla_{\mathcal{G}^t} \log q^{t}\left(\mathcal{G}^t\right) \approx - \frac{\boldsymbol{\hat{\epsilon}}}{\sqrt{1-\bar{\alpha}^t}}$. Thus, the gradient in~\cref{eq:smgradient} can be written as
\begin{equation}
\label{eq:gradient_calculate}
\begin{aligned}
    & \nabla_\eta D_{\mathrm{KL}} \left(p_G^{t} \left( \mathcal{G}^t \right) \| q^{t}\left(\mathcal{G}^t\right)\right) \\ & = 
    \mathbb{E}_{t,\epsilon} \left[ \nabla_{\mathcal{G}^t}\log p_G^{t}\left(\mathcal{G}^t\right) - \nabla_{\mathcal{G}^t} \log q^{t}\left(\mathcal{G}^t\right)\right] \frac{\partial \mathcal{G}^t}{\partial \eta} \\ & \approx
    \mathbb{E}_{t,\epsilon} \left[-\frac{\boldsymbol{\epsilon}_{\phi}\left(\mathcal{G}^{t}, \mathcal{P}, t\right)}{\sqrt{1-\bar{\alpha}^t}} - \left(-\frac{\boldsymbol{\epsilon}_{\theta}\left(\mathcal{G}^{t}, \mathcal{P}, t\right)}{\sqrt{1-\bar{\alpha}^t}}\right)\right] \frac{\partial \mathcal{G}^t}{\partial \eta} \\ & =
    \mathbb{E}_{t,\epsilon} \left[\frac{\boldsymbol{\epsilon}_{\theta}\left(\mathcal{G}^{t}, \mathcal{P}, t\right)}{\sqrt{1-\bar{\alpha}^t}} - \frac{\boldsymbol{\epsilon}_{\phi}\left(\mathcal{G}^{t}, \mathcal{P}, t\right)}{\sqrt{1-\bar{\alpha}^t}}\right] \frac{\partial \mathcal{G}^t}{\partial \eta}
\end{aligned}
\end{equation}
Here $\sqrt{1-\bar{\alpha}^t}$ can be ignored. Thus, the gradient of $G_{stu}$ can be approximated by 
\begin{equation}
\label{eq:smG_gradient}
\begin{aligned}
    & \nabla_\eta D_{\mathrm{KL}} \left(p_G^{t} \left( \mathcal{G}^t \right) \| q^{t}\left(\mathcal{G}^t\right)\right) \\ & \approx 
    \mathbb{E}_{t,\epsilon}\left[ \boldsymbol{\epsilon}_{\theta}\left(\mathcal{G}^{t}, \mathcal{P}, t\right)- \boldsymbol{\epsilon}_{\phi}\left(\mathcal{G}^{t}, \mathcal{P}, t\right)\right] \frac{\partial \mathcal{G}^t}{\partial \eta} 
\end{aligned}
\end{equation}

\section{Introduction on utilized methods}
\subsection{Variational score distillation}
Variational Score Distillation (VSD), proposed by ProlificDreamer~\cite{prolificdreamer}, is designed to leverage a pre-trained diffusion model to train a NeRF~\cite{nerf}, enabling the rendering of high-quality 3D objects. 

Given a text prompt $y$, the probabilistic distribution of all possible 3D representations can be modeled as a probabilistic density $\mu(\theta\|y)$ by a NeRF model parameterized by $\theta$. Let $q_0^\mu(\boldsymbol{x}_0\|c, y)$ as the distribution of the rendered image $\boldsymbol{x}_0$ of NeRF given the camera $c$, and $p_0(\boldsymbol{x}_0\|y)$ as the distribution of the pre-trained text-to-image diffusion model at $t=0$. To generate high-quality 3D objects, ProlificDreamer~\cite{prolificdreamer} optimizes the distribution of $\mu$ by minimizing the following KL divergence
\begin{equation}
    \min _{\mu} D_{\mathrm{KL}}\left(q_{0}^{\mu}\left(\boldsymbol{x}_{0} \mid y\right) \| p_{0}\left(\boldsymbol{x}_{0} \mid y\right)\right)
    \label{eq:pd}
\end{equation}

However, directly solving this variational inference problem is challenging because $p_0$ is complex, and its high-density regions may be extremely sparse in high-dimensional spaces. Therefore, ProlificDreamer reformulates it as an optimization problem at different time steps $t$, referring to these problems as Variational Score Distillation (VSD),
\begin{equation}
    \min _{\mu} \mathbb{E}_{t,c}\left[\left(\sigma_{t} / \alpha_{t}\right) \omega(t) D_{\mathrm{KL}}\left(q_{t}^{\mu}\left(\boldsymbol{x}_{t} \mid c,y\right) \| p_{t}\left(\boldsymbol{x}_{t} \mid y\right)\right)\right]
    \label{eq:vsd}
\end{equation}

Theorem 1 in~\cite{prolificdreamer} proves that introducing the additional $t$ does not affect the global optimum of~\cref{eq:pd}. Theorem 2 in~\cite{prolificdreamer} provides the method for optimizing the problem in~\cref{eq:vsd}.
\begin{equation}
\begin{aligned}
    \frac{\mathrm{d} \theta_{\tau}}{\mathrm{d} \tau} &=-\mathbb{E}_{t, \epsilon, c}[\omega(t)(\underbrace{-\sigma_{t} \nabla_{\boldsymbol{x}_{t}} \log p_{t}\left(\boldsymbol{x}_{t} \mid y\right)}_{\text {score of noisy real images }} \\ &-\underbrace{\left(-\sigma_{t} \nabla_{\boldsymbol{x}_{t}} \log q_{t}^{\mu_{\tau}}\left(\boldsymbol{x}_{t} \mid c, y\right)\right)}_{\text {score of noisy rendered images }}) \frac{\partial \boldsymbol{g}\left(\theta_{\tau}, c\right)}{\partial \theta_{\tau}}]
\end{aligned}
\end{equation}
Here the score of noisy real images is approximated by the pre-trained diffusion model $\boldsymbol{\epsilon}_{pretrain}(\boldsymbol{x}_{t},t,y)$ and the score of noisy rendered images is approximated by another diffusion model $\boldsymbol{\epsilon}_{\phi}(\boldsymbol{x}_{t},t,c,y)$, which is trained on the rendered images with the standard diffusion objective. 
\begin{equation}
    \min _{\phi} \sum_{i=1}^{n} \mathbb{E}_{t, \epsilon, c }\left[\left\|\boldsymbol{\epsilon}_{\phi}\left(\alpha_{t} \boldsymbol{g}\left(\theta^{(i)}, c\right)+\sigma_{t} \boldsymbol{\epsilon}, t, c, y\right)-\boldsymbol{\epsilon}\right\|_{2}^{2}\right]
\end{equation}

In practice, $\boldsymbol{\epsilon}_{\phi}(\boldsymbol{x}_{t},t,c,y)$ is parameterized by a small UNet or the Low-rank adaptation (LoRA)~\cite{LoRA} of the teacher model. With the alternating training of NeRF and $\boldsymbol{\epsilon}_{\phi}(\boldsymbol{x}_{t},t,c,y)$, ProlificDreamer~\cite{prolificdreamer} is ultimately able to generate high-quality 3D objects.

\subsection{MinkowskiEngine}
Sparse tensor computation plays a critical role in fields such as 3D point cloud processing, computer vision, and physical simulations. Unlike dense tensors, sparse tensors contain a high proportion of zero values and directly applying traditional tensor operations can lead to inefficient use of computational resources. Minkowski Engine~\cite{Minkowski} addresses these challenges by providing a high-performance framework tailored for sparse tensor computation, enabling efficient operations on high-dimensional sparse data. In this paper, we used the Minkowski Engine to process sparse point cloud data.

Minkowski Engine introduces several innovative approaches to sparse tensor processing. 
\begin{itemize}
    \item Efficient Sparse Tensor Representation. 
    Sparse tensors are represented using coordinate-value pairs, eliminating the need to store zeros. This representation reduces both memory usage and computational overhead.
    \item Sparse Convolution Operations
    The framework supports high-dimensional sparse convolutions, with kernels designed to adapt to varying sparsity patterns. Optimized memory access patterns and parallel computation strategies ensure high efficiency.
    \item Fast Coordinate Mapping
    Minkowski Engine employs hash tables for rapid coordinate mapping, which accelerates tensor indexing and sparse pattern matching.
    \item Automatic Differentiation Support
    The framework includes built-in support for automatic differentiation, facilitating the training of machine learning models based on sparse tensors.
    \item Multi-Dimensional Capability
    Minkowski Engine can handle sparse tensors of arbitrary dimensions, making it suitable for a wide range of applications, from 2D image processing to 5D simulations.
\end{itemize}
Minkowski Engine has been widely adopted in various domains including 3D point cloud processing, physical simulations and medical imaging. By significantly improving computational efficiency and scalability, the Minkowski Engine has become a preferred choice for handling sparse tensor computations in both research and industrial applications. 
\section{Ethical statement}
The potential ethical impact of our work is about fairness.
As ``human'' is included as a kind of object in the LiDAR scene, when performing scene completion, it may be necessary to complete human figures. Human-related objects may have data bias related to fairness issues, such as the bias to gender or skin colour. Such bias can be captured by the student model in the training. 

\subsection{Notification to human subjects}

In our user study, we present the notification to subjects to inform the collection and use of data before the experiments.

\begin{quote}
    
Dear volunteers, we would like to thank you for supporting our study. We propose ScoreLiDAR, a novel distillation method tailored for 3D LiDAR scene completion, which introduces a structural loss to help the student model capture the geometric structure information. All information about your participation in the study will appear in the study record. All information will be processed and stored according to the local law and policy on privacy. Your name will not appear in the final report. Only an individual number assigned to you is mentioned when referring to the data you provided.

We respect your decision whether you want to be a volunteer for the study. If you decide to participate in the study, you can sign this informed consent form.
\end{quote}

The Institutional Review Board approved the use of users' data of the main authors' affiliation.
\section{Failure examples}
\begin{figure*}
    \centering
    \includegraphics[width=\linewidth]{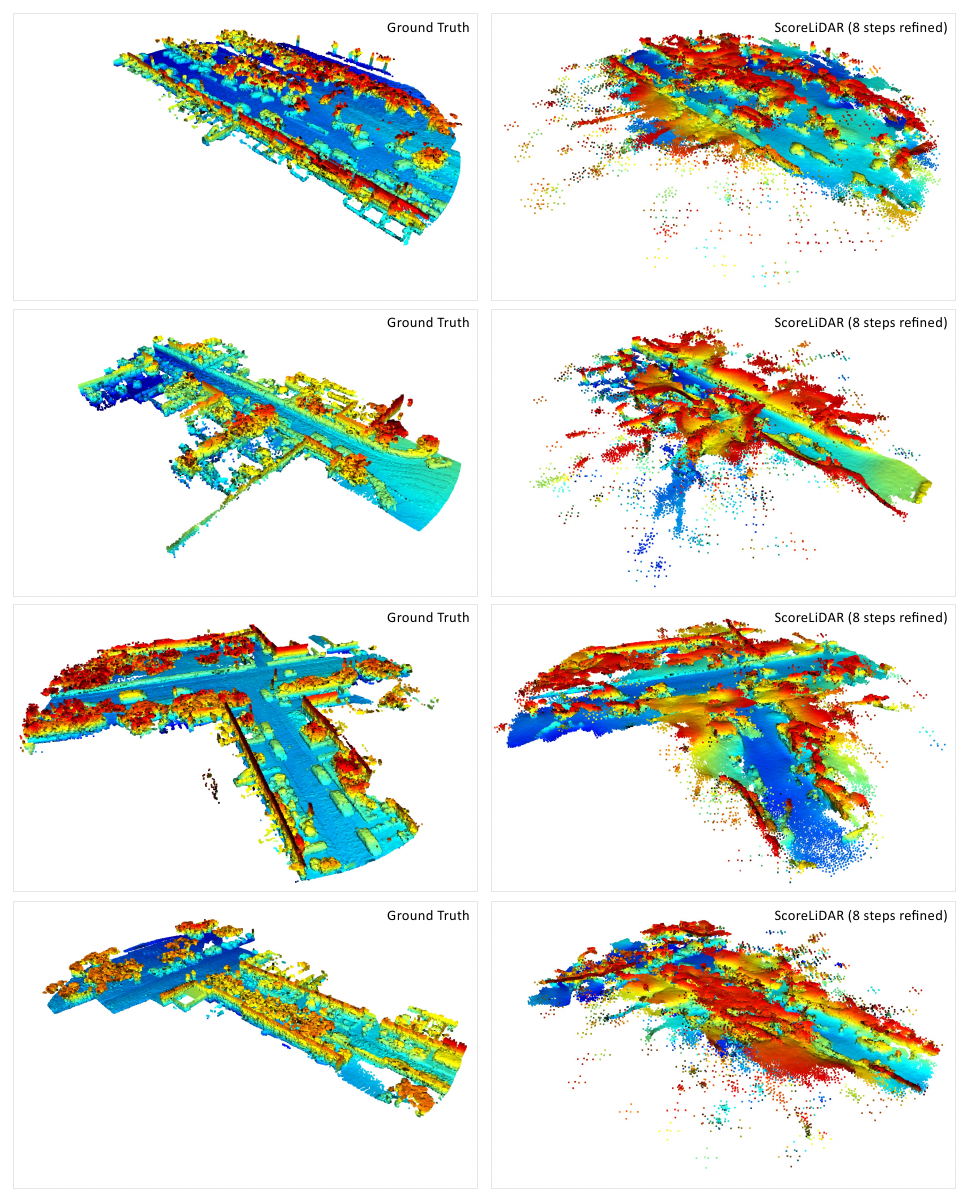}
    \caption{Failure examples of ScoreLiDAR.}
    \label{fig:failure}
\end{figure*}

\cref{fig:failure} presents some failure cases of ScoreLiDAR. From these examples, it can be observed that ScoreLiDAR exhibits over-completion to some extent, where regions that do not exist are completed. Before the completion, as mentioned in Sec.3 in the main paper, the number of points of the input sparse scan $\mathcal{P}$ is increased by concatenating its points $K$ times and the dense input $\mathcal{P}^*$ is obtained. If the number of points of $\mathcal{P}^*$ exceeds the actual number of points in the ground truth, it can lead to redundant points in the completed scene. These redundant points may be distributed in areas that do not require completion, resulting in the situations observed in the failure cases.

\end{document}